\documentclass[10pt,twocolumn,letterpaper]{article}

\usepackage{iccv}
\usepackage{times}
\usepackage{epsfig}
\usepackage{graphicx}
\usepackage{amsmath}
\usepackage{amssymb}
\usepackage{booktabs}
\usepackage{algorithmic}
\usepackage{algorithm}
\usepackage{color}
\usepackage{multirow}
\usepackage{booktabs}
\usepackage{pifont}
\usepackage{caption}
\usepackage[pagebackref=true,breaklinks=true,letterpaper=true,colorlinks,bookmarks=false]{hyperref}

\iccvfinalcopy 

\begin{document}
\captionsetup[figure]{labelfont={small},labelformat={default},labelsep=period,name={Figure}}
\captionsetup[table]{labelfont={small},labelformat={default},labelsep=period,name={Table}}
\title{Progressive Learning with Cross-Window Consistency for Semi-Supervised Semantic Segmentation}

\author{Bo Dang \qquad Yansheng Li \thanks{corresponding author.} \qquad Yongjun Zhang \qquad Jiayi Ma\\
Wuhan University, Wuhan, China\\
{\tt\small \{bodang, yansheng.li, zhangyj\}@whu.edu.cn~~jyma2010@gmail.com}}

\maketitle

\begin{abstract}
   Semi-supervised semantic segmentation focuses on the exploration of a small amount of labeled data and a large amount of unlabeled data, which is more in line with the demands of real-world image understanding applications. However, it is still hindered by the inability to fully and effectively leverage unlabeled images. In this paper, we reveal that \textbf{c}ross-\textbf{w}indow \textbf{c}onsistency (CWC) is helpful in comprehensively extracting auxiliary supervision from unlabeled data. Additionally, we propose a novel CWC-driven progressive learning framework to optimize the deep network by mining weak-to-strong constraints from massive unlabeled data. More specifically, this paper presents a \textbf{b}iased \textbf{c}ross-window \textbf{c}onsistency (BCC) loss with an importance factor, which helps the deep network explicitly constrain confidence maps from overlapping regions in different windows to maintain semantic consistency with larger contexts. In addition, we propose a \textbf{d}ynamic \textbf{p}seudo-label \textbf{m}emory bank (DPM) to provide high-consistency and high-reliability pseudo-labels to further optimize the network. Extensive experiments on three representative datasets of urban views, medical scenarios, and satellite scenes demonstrate our framework consistently outperforms the state-of-the-art methods with a large margin. Code will be available publicly.
\end{abstract}

\begin{figure}
\centering
\includegraphics[scale=.225]{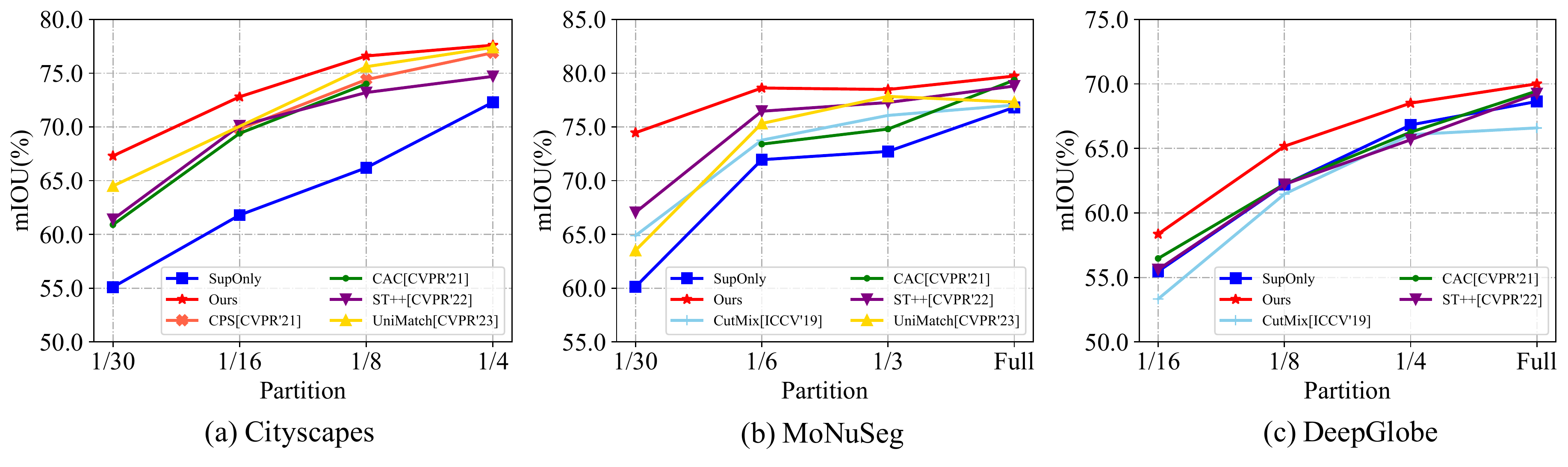}
\vspace{-2.0em}
\caption{Performance (mean intersection-over-union, mIOU) comparisons between our framework and the state-of-the-art methods on various datasets from multiple fields: (a) urban street scenes segmentation (\ie~, Cityscapes \cite{cordts2016cityscapes}), (b) medical nuclear segmentation (\ie~, MoNuSeg \cite{kumar2019multi}), and (c) land cover classification (\ie~, DeepGlobe \cite{demir2018deepglobe}). It is worth noting that our method (\textcolor{red}{red lines}) surpasses existing methods significantly on various datasets under different partition protocols. Methods studied: SupOnly (supervised baseline), CutMix \cite{yun2019cutmix}, CPS \cite{chen2021semi}, CAC \cite{lai2021semi}, ST++ \cite{yang2022st++}, UniMatch \cite{unimatch}, and the proposed framework. See Section~\ref{sec:sec4} for experiment details.}
\vspace{-1.5em}
\label{fig:fig1} 
\end{figure}

\section{Introduction}
\label{sec:intro}

Semantic segmentation, as a fundamental and essential task, is widely employed in a wide range of situations, such as automated driving, medical pathology diagnosis, and land cover survey \cite{zhao2017pyramid, ronneberger2015u, tong2020land}. The brilliant performance of data-driven deep learning algorithms largely depends on huge volumes of annotated data. In practice, massive unlabeled images are collected, but it is hard to acquire the corresponding pixel-level annotations. Despite the availability of advanced semi-automatic labeling algorithms \cite{ling2019fast,hao2021edgeflow}, the process of generating annotated data is still tremendously labor-intensive and time-consuming, particularly the annotating process of remote sensing and medical images requires the participation of experts with domain knowledge. To alleviate this issue, numerous semi-supervised learning methods \cite{souly2017semi,ouali2020semi,chen2021semi,yang2022st++,liu2022perturbed,fan2022ucc,wu2022cross,sun2020bas} have been developed and achieve promising performance.

Although lots of achievements have been obtained in semi-supervised semantic segmentation, many tricky challenges still remain. The first challenge is how to generate or select pseudo-labels with high-reliability for preventing catastrophic performance degradation. Minimizing the adverse impact of the noise of pseudo-labels is a longstanding but unsolved issue in self-training pipelines \cite{yang2022st++}. The second challenge is that heterogeneous consistency traits are not fully utilized. Contextual-aware consistency \cite{lai2021semi} can be viewed as a unique form of data augmentation (\ie~, contextual augmentation) and applied to unlabeled data. Similar ideas are also involved in self-supervised learning \cite{chen2021multisiam} and image-to-image translation \cite{ko2022self}, which shows that \textbf{c}ross-\textbf{w}indow \textbf{c}onsistency (CWC) is promising. However, it is still insufficient for existing works to fully exploit the merit of CWC. For instance, Directional Contrastive loss from \cite{lai2021semi} requires manual setting of some key parameters (such as the positive filtering threshold) that must be tuned depending on the datasets. As a whole, CWC is preliminarily explored in the consistency loss modeling, but unfortunately ignored in the selection of high-quality pseudo-labels.

Similar to the peripheral vision system in human vision \cite{lettvin1976seeing,min2022peri}, human visual reasoning processes need to rely on multiple contour regions that cover different contextual information. In life, when humans view images from cross windows, the visual center of the cerebral cortex often produces the same response on overlapping regions. These facts guide us to leverage CWC to exploit unlabeled data.

In light of the aforementioned challenges and the inspiration of human vision, we propose a progressive learning framework on the fundamental idea that \textbf{overlapping regions on image patches from diverse contextual windows exhibit semantic consistency}, systematically exploiting the benefits of this inherent consistency. Our framework progressively optimizes deep network by mining weak-to-strong constraints from unlabeled data. Specifically, in the first stage, we introduce a general and effective \textbf{b}iased \textbf{c}ross-window \textbf{c}onsistency (BCC) loss that measures the semantic consistency of overlapping regions based on the segmentation confidence maps. In the second stage, we further extend this fundamental concept by designing a unique pseudo-label reliability evaluating method and establishing a highly dynamic and rewarding \textbf{d}ynamic \textbf{p}seudo-label \textbf{m}emory bank (DPM) to assist in exposing the model to strong pseudo-label constraints. Benefiting from our proposed pseudo-label reliability evaluation algorithm guided by the inherent cross-window dependencies of images and a well-designed DPM, our approach does not compute the pseudo-label discrepancy of the model at multiple phases as in ST++ \cite{yang2022st++}, but instead only computes the contextual prediction consistency of overlapping regions in different windows to to ensure the information in the DPM is constantly and dynamically updated.

Our framework is generalized and can be extended simply to semi-supervised semantic segmentation applications (\eg~, urban street scenes segmentation in computer vision, medical nuclear segmentation in pathological analysis, and land cover classification in remote sensing). Extensive experiments on the Cityscapes \cite{cordts2016cityscapes}, MoNuSeg \cite{kumar2019multi}, and DeepGlobe \cite{demir2018deepglobe} datasets demonstrate a considerable performance improvement over the state-of-the-art methods, as shown in Figure~\ref{fig:fig1}. By systematically exploring CWC, our main contributions are summarized as follows:
\begin{itemize}
  \vspace{-0.5em}
  \item The BCC loss with the importance factor is designed to maintain larger contextual semantic consistency among overlapping confidence maps.
  \vspace{-0.5em}
  \item We propose a DPM using a novel pseudo-label reliability evaluation method to minimize the adverse effects of ill-posed pseudo-labels.
  \vspace{-0.5em}
  \item Our framework outperforms previous methods on extensive datasets from different fields, which demonstrates the strong generalization and competitiveness of our work.
\end{itemize}

\section{Related Work}
\noindent \textbf{Semi-supervised semantic segmentation}‘s crux and core is how to properly utilize unlabeled data and be able to further enhance the generalization of the model with less labeled data. With the rapid improvement of semi-supervised learning (SSL) methods \cite{sohn2020fixmatch,grandvalet2004semi,zoph2020rethinking,miyato2018virtual,berthelot2019mixmatch}, solutions based on different paradigms have made progress in semi-supervised semantic segmentation tasks. The current semi-supervised semantic segmentation approach consists of three typical pipelines: GAN-based, self-training, and consistency regularization methods. In semi-supervised semantic segmentation, GANs \cite{goodfellow2020generative,mittal2019semi} are used as discriminative tools or supervised signals alone or in conjunction with other methods. For example, Hung \etal~\cite{hung2018adversarial} uses discriminators of GAN networks to find pseudolabeled plausible regions. Previous works \cite{mittal2019semi,hou2022semi} add the GAN branch as an auxilary supervision in natural and medical images, respectively. 

\noindent \textbf{Consistency regularization-based methods} make features of samples from the same category more compact in the feature space, while keeping features of samples from different categories as far as possible. The benefit comes in the implement's flexibility, which includes the design and metrics of consistency traits. Specifically, CutMix \cite{yun2019cutmix}, ClassMix \cite{olsson2021classmix}, and various other data augmentations \cite{shorten2019survey} are federated in the consistency regularization framework in order to transform or perturb the input data to satisfy the constraints of the consistency measure, just as \cite{ouali2020semi,liu2022perturbed} do. Similarly, further broader disturbances and different initialization model are published to achieve a gain \cite{fan2022ucc,chen2021semi}, respectively. Contrastive loss that performs well on other tasks is relocated to the consistency regularization paradigm in owing to the rapid advancement of contrastive learning and self-supervised learning \cite{ye2019unsupervised,jaiswal2020survey,he2020momentum,chen2021exploring}. For instance, InfoNCE \cite{oord2018representation}, which attempts to bring positive pairs closer and push negative pairs apart and shines in self-supervised learning, has been extensively modified and adapted to many previous methods \cite{zhong2021pixel,xu2022semi,lai2021semi,wu2022cross,xiao2022semi}. In addition, CCT \cite{ouali2020semi} emphasizes the validity of the mean square error (MSE) as an elegant consistency loss.

\noindent \textbf{Self-training-based and Pseudo labeling-based methods} commonly leverage student-teacher models to produce and re-train pseudo-labels. Chronic challenges include how to generate or select pseudo-labels with high-confidence to optimize the model and tackle the class-imbalanced issue. In response to the first problem outlined, ST++ \cite{yang2022st++} proposes a straightforward yet effective pipeline that boosts model stability through strong and weak data transformations and by gradually utilizing all pseudo-labels. Instead of ignoring the doubtful pixels of pseudo-labels, U$^{2}$PL \cite{wang2022semi} treats them as negative samples to be compared with the matching positive samples. ELN \cite{kwon2022semi} and Yuan \etal~\cite{yuan2021simple} create the ELN module to correct pseudo labels and self-correction loss to prevent overfiting to the noise of low-confidence pseudo-labels, respectively. The class-imbalance bias of pseudo-labels undermine the generalization of the model, particularly when there are very few unlabeled samples or when the sample contains a significant long-tail effect. Numerous solutions \cite{guan2022unbiased,hu2021semi,he2021re} recognize this issue and align class distributions to rectify the imbalance. Note that the existing methods do not perfectly address the mentioned challenges.

Different from previous methods, our framework focuses on exploring CWC and brings significant performance gain by minimizing differences among unlabeled data across diverse windows to mitigate cross-window bias and dynamically selecting rewarding pseudo-labels to avoid the misleading of ill-posed pseudo-labels and overfitting of fixed pseudo-labels.

\section{Method}
\subsection{Problem Definition}
The goal of semi-supervised semantic segmentation is to employ a small set of labeled data $ \mathcal{B}_l=\left\{\left(x_i, y_i\right)\right\}_{i=1}^M $ with unlabeled data $ \mathcal{B}_u=\left\{u_i\right\}_{i=1}^N $ to train a model $ \mathcal{F} $ that can provide accurate results on test data. In general, the overall optimization loss can be formulated as:
\vspace{-0.2em}
\begin{align}
\label{deqn_ex1}
\mathcal{L}_{total}=\mathcal{L}_s+\lambda \mathcal{L}_u,
\end{align}
where $ \lambda $ is a trade-off weight between labeled and unlabeled data supervision. Typically, the labeled supervised loss $ \mathcal{L}_s $ is the cross-entropy loss or correlation variant (\eg~, OHEM \cite{shrivastava2016training}) of the inferences and annotated labels. The unsupervised loss $ \mathcal{L}_u $ can be defined flexibly as consistency loss, pseudo-label loss, entropy minimum loss, \etc~, thereby encouraging the model to fit the unlabeled data.

\subsection{Motivation and Overview}
\begin{figure}
\centering
\includegraphics[scale=.35]{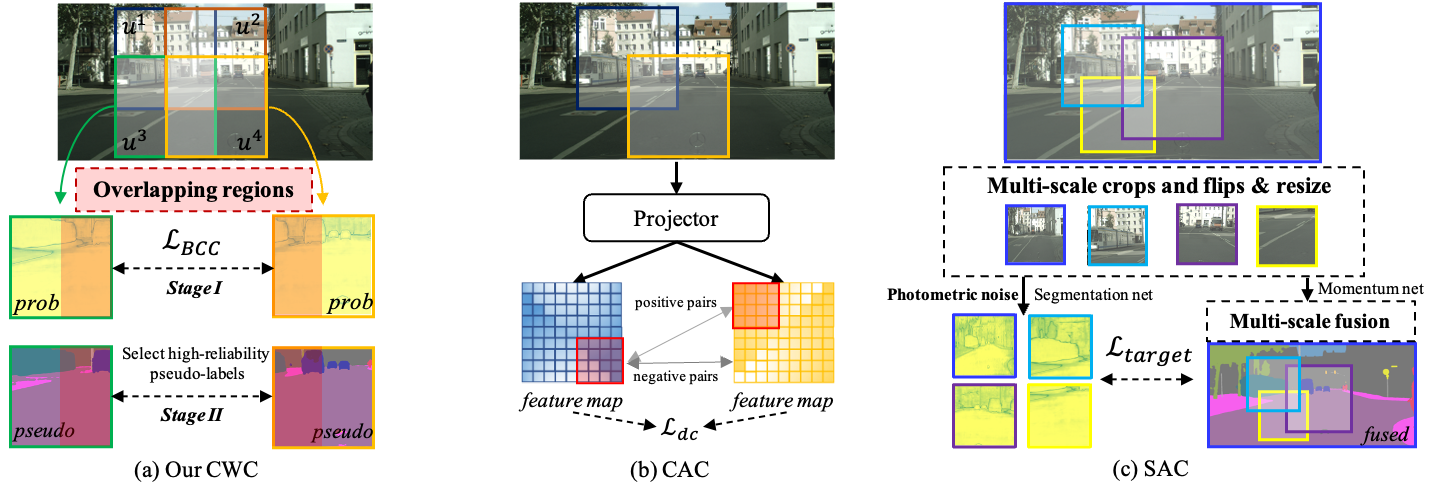}
\vspace{-2.0em}
\caption{Emphasize the differences between our approach and previous work. In contrast to (a) CAC \cite{lai2021semi}, we employ the stable confidence maps rather than feature maps to encourage consistency of overlapping regions from wider contextual windows, and further apply CWC traits to select high-reliability pseudo-labels. (c) SAC \cite{Araslanov_2021_CVPR}, which is designed for domain adaptation, is to enforce consistency of the semantic maps created by the model across image perturbations rather than utilizing the consistency of overlapping regions across windows as we do.}
\vspace{-1.5em}
\label{fig:fig6}
\end{figure}

The majority of existing consistency regularization-based approaches \cite{ouali2020semi,zhong2021pixel,xiao2022semi,xu2022semi} focus on learning feature consistency following perturbation or data augmentation, whereas CAC \cite{lai2021semi} introduces a context-aware consistency loss that compares high-level feature consistency. In contrast to it, (1) we focus on the fact that the feature representation computation is unstable (even with the participation of the feature projection \cite{chen2020simple}), and (2) we apply the spontaneous existing image CWC traits to filter pseudo-labels, and subsequent experiments demonstrate its effectiveness, as highlighted in Figure~\ref{fig:fig6}.

Motivated by the aforesaid discussions, our overall optimization objective function can be defined as follows:
\begin{align}
\label{deqn_ex2}
\mathcal{L}_{total}=\mathcal{L}_s+\lambda_{BCC} \mathcal{L}_{BCC}+\lambda_{DPM} \mathcal{L}_{DPM},
\end{align}
\vspace{-2.5em}
\begin{align}
\label{deqn_ex3}
\mathcal{L}_s=\frac{1}{\left|\mathcal{B}_l\right|} \sum\limits_{\mathbf{X} \in \mathcal{B}_l} \frac{1}{W \times H} \sum\limits_{i=0}^{W \times H}\left(\ell_{c e}\left(\mathbf{p}_{i}, y_{i}\right)\right),
\end{align}
where $W$ and $H$ represent the width and height of images. For the labeled dataset $ \mathcal{B}_l $, the semantic segmentation model $ \mathcal{F} $ is employed to generate its confidence map $\mathbf{p}_{i}$ (after $Softmax$ normalization), which is supervised by the ground truth $y_{i}$ using the cross-entropy loss $\ell_{c e}$ . As for the unlabeled dataset $\mathcal{B}_u$, our BCC loss $\mathcal{L}_{BCC}$ reflects the weak constraint, as described in Section~\ref{sec:stage 1}. The pseudo-label supervised loss $\mathcal{L}_{DPM}$ is the strong constraint, and Section~\ref{sec:stage 2} describes the pseudo-label filtering and usage.  

\begin{figure*}
	\centering
		\includegraphics[scale=.45]{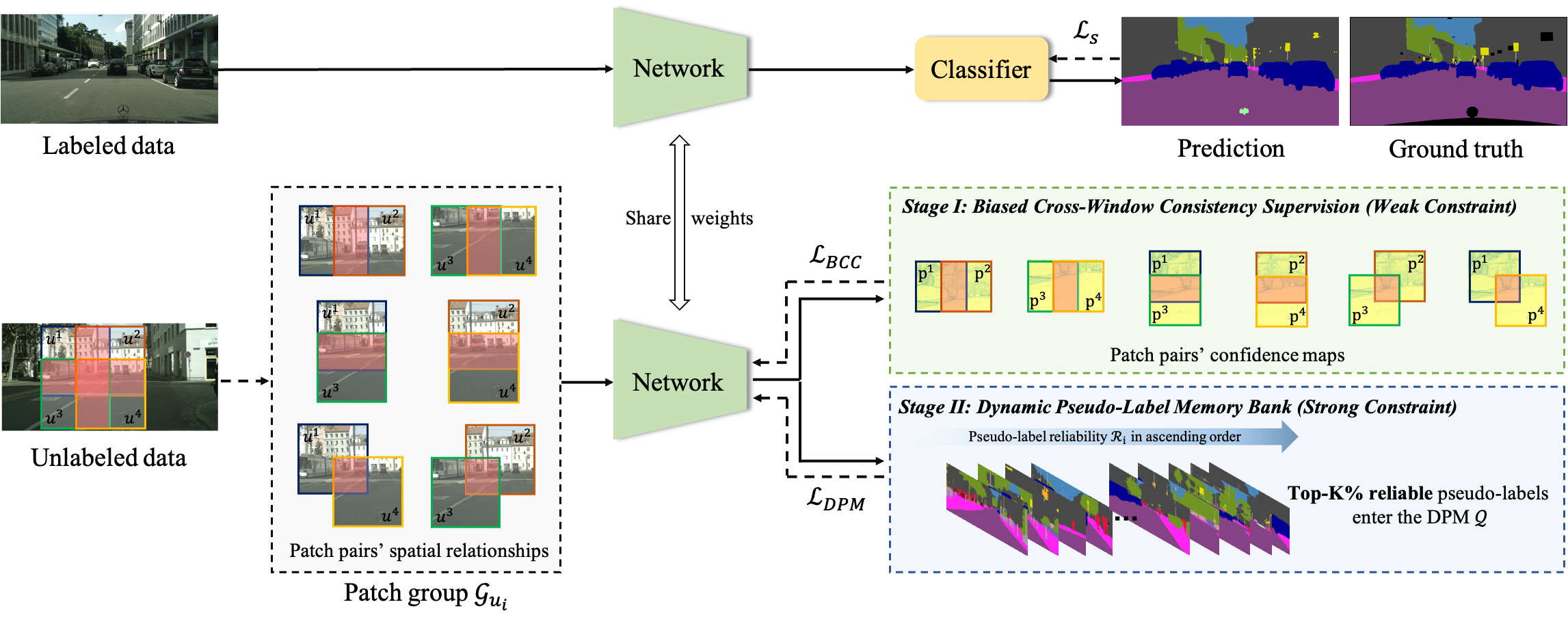}
            \vspace{-2.0em}
	    \caption{An overview of our proposed framework. labeled data is employed for supervised training the model $\mathcal{F}$. For unlabeled data, in stage I, we present a novel BCC loss with an importance factor (Eq.~\eqref{con:deqn_ex4}) to encourages the model to maintain consistency across overlapping confidence maps in different windows but does not restrict specific class attributes. In addition, we propose the DPM to rank the pseudo-label reliability (Eqs.~\eqref{con:deqn_ex7} and \eqref{con:deqn_ex8}) in light of CWC and dynamically update and manage rewarding pseudo-labels.}
	\label{fig:fig2}
 \vspace{-1.5em}
\end{figure*}

Ideally, all items of the Eq.~\eqref{deqn_ex2} are optimized together, but it is extremely GPU memory-consuming. To operate pervasively on the overwhelming majority of devices, we propose a progressive learning strategy to achieve the ultimate optimization goal. Algorithm~\ref{alg:AOS} and Figure~\ref{fig:fig2} offer a comprehensive pseudocode description and an intuitive overivew of our whole framework, respectively.

\begin{algorithm}[!h]
    \caption{Our Framework Pseudocode}
    \label{alg:AOS}
    \renewcommand{\algorithmicrequire}{\textbf{Input:}}
    \renewcommand{\algorithmicensure}{\textbf{Output:}}
    
    \begin{algorithmic}[1]
        \REQUIRE Labeled images and corresponding labels $ \mathcal{B}_l=\left\{\left(x_i, y_i\right)\right\}_{i=1}^M $ \\ 
                 Unlabeled images $\mathcal{B}_u=\left\{u_i\right\}_{i=1}^N $\\
                 Validation set $ \mathcal{B}_v=\left\{\left(x_i, y_i\right)\right\}_{i=1}^V $
        \ENSURE Trained model $\mathcal{F}$    
        
        \STATE  \textcolor{blue}{\#Stage I: Biased Cross-Window consistency supervision (weak constraint)}
        \STATE Train $\mathcal{F}$ on $\mathcal{B}_{l}$ with cross-entropy loss and $\mathcal{B}_{u}$ based on Eq.~\eqref{con:deqn_ex4}
        \STATE Initialize previous best $\mathcal{S}\leftarrow0$
        \STATE Calculate previous best $\mathcal{S}\leftarrow \operatorname{meanIOU}\left(\mathcal{F}\left(\mathcal{B}_{v}\right)\right)$
        \STATE Initialize count $count\leftarrow0$
        \WHILE{epoch$<$maximum number of epochs}
            \STATE  \textcolor{blue}{\#Stage II: Dynamic pseudo-label memory bank (strong constraint)}
            \FOR{$u_{i}$ $\in$ $\mathcal{B}_{u}$}
                \STATE Get reliability score $\mathcal{R}_{i}$ based on Eqs.~\eqref{con:deqn_ex7} and \eqref{con:deqn_ex8}
            \ENDFOR
            \STATE Select Top-K\% scored unlabeled images and generate corresponding pseudo-labels $y_{i}^*$ into the dynamic pseudo-label memory bank $\mathcal{Q}=\left\{\left(u_i, y_{i}^{*}\right)\right\}_{i=1}^{N\times K\%}$
            \STATE Train $\mathcal{F}$ on $\mathcal{B}_{l}\cup\mathcal{Q}$ with cross-entropy loss
            \STATE $count\leftarrow count+1$
            \STATE Calculate current $\mathcal{S}^{'}\leftarrow \operatorname{meanIOU}\left(\mathcal{F}\left(\mathcal{B}_{v}\right)\right)$
                \IF {$ \mathcal{S}^{'} > \mathcal{S} + \mathcal{G}$ \OR $count > \mathcal{K}$}
                    \STATE Update the $\mathcal{Q}$ (Jump to Stage II)
                    \STATE Update previous best $\mathcal{S}\leftarrow\mathcal{S^{'}}$
                    \STATE Re-initialize count $count\leftarrow 0$
                \ENDIF
        \ENDWHILE
        
        \RETURN $\mathcal{F}$
    \end{algorithmic}
\end{algorithm}

\subsection{Stage I: Biased Cross-Window Consistency Supervision (Weak Constraint)}
\label{sec:stage 1}
For each unlabeled image $u_i$, four adjacent and overlapping patches (default minimum overlap size is $\frac{H}{2}$ or $\frac{W}{2}$) are defined as a patch group $ \mathcal{G}_{u_ i}=\left\{\left(u_i^1, u_i^2, u_i^3, u_i^4\right)\right\}_{i=1}^N $, and their spatial relationships are depicted in Figure~\ref{fig:fig2}. Any patch pairs $ \left(u_i^k, u_i^{l}\right) $ containing overlapping regions $ u_{oi} $ are processed by the encoder $\mathcal{E}$, decoder $\mathcal{D}$, and classifier $\mathcal{C}$ to obtain a $softmax$ normalized confidence map $\mathbf{p}_{i}$, which is then used to calculate the BCC loss. 

\begin{figure}
\centering
\includegraphics[scale=.35]{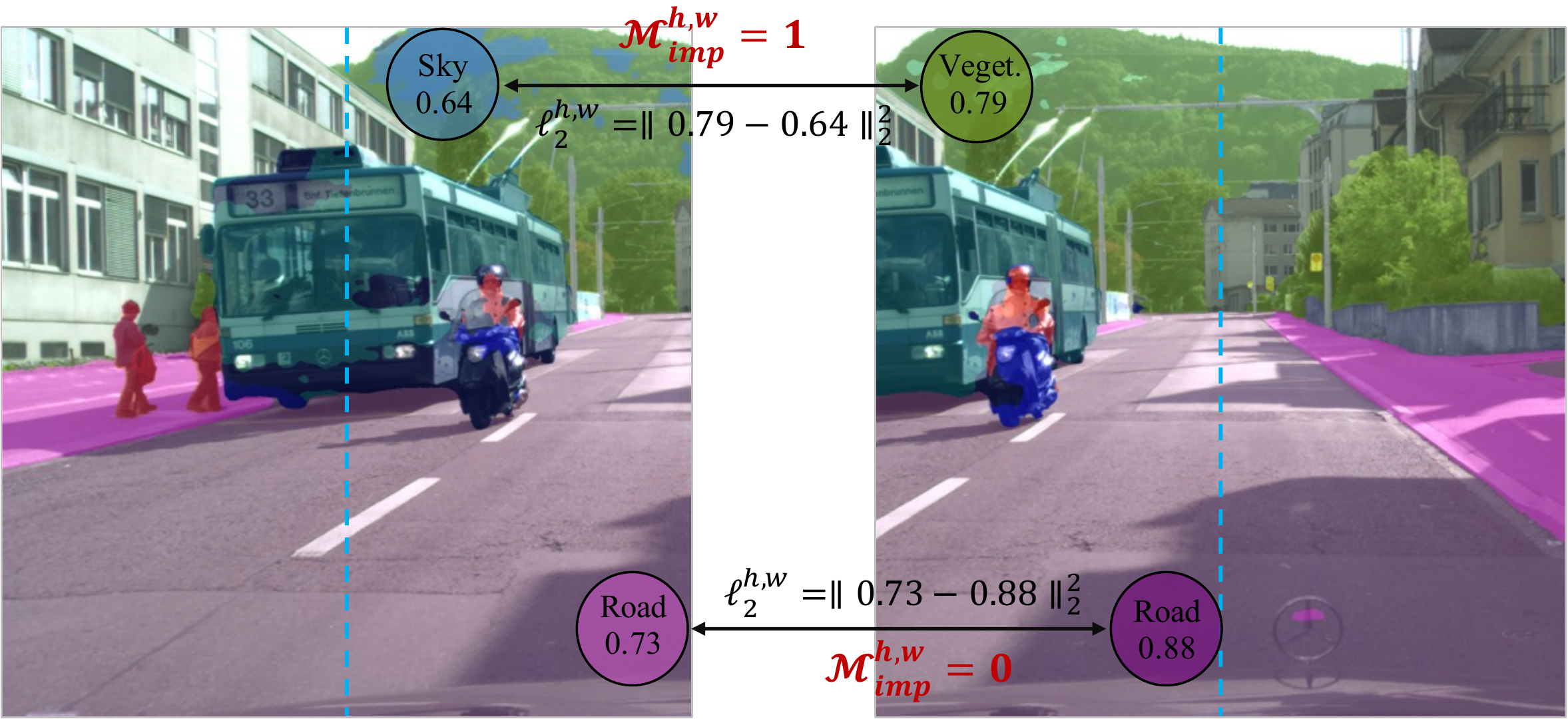}
\vspace{-2.0em}
\caption{Illustration of the importance factor $\mathcal{M}_{imp}^{h,w}$. It eliminates insignificant differences (pixel positions with different confidence maps but the same semantic classes)}
\vspace{-1.5em}
\label{fig:fig3}
\end{figure}

BCC loss encourages overlapping regions within the pair groups to have semantically consistent representations but does not restrict specific class attributes. In particular, to encourage the model to focus on \textbf{prominent} differences with different semantic classes in overlapping regions, we propose the importance factor $\mathcal{M}_{imp}^{h,w}$ to eliminate insignificant differences (pixel positions with different confidence maps but the same semantic classes), as illustrated in Figure~\ref{fig:fig3}. Intuitively, the importance factor $\mathcal{M}_{imp}^{h,w}$ will amplify the difference between the feature information of pixels that are more valuable to the model. Table~\ref{tab:table4} experimental results demonstrate its benefits. Our BCC loss can be written as
\begin{align}
\begin{split}
\label{con:deqn_ex4}
&\mathcal{L}_{BCC}= \frac{1}{\left|B_u\right|} \sum_{x \in B_u} \\
&\frac{1}{W \times H} \sum_{i=0}^{W \times H} \sum_{1 \leqslant k<l \leqslant 4} \ell_{2}\left(\mathbf{p}_{o i}^k, \mathbf{p}_{o i}^{l}\right) \cdot \mathcal{M}_{imp}^{h,w},
\end{split}
\end{align}
\vspace{-1.5em}
\begin{align}
\label{deqn_ex5}
\mathbf{p}_{o i}^k=\operatorname{Softmax}\left(\mathcal{C}\left(u_{o i}^k\right)\right),
\end{align}
\vspace{-2.5em}
\begin{equation}
\setlength\abovedisplayskip{2pt}
\setlength\belowdisplayskip{1pt}
\begin{split}
\label{deqn_ex6}
&\mathcal{M}_{i m p}^{h, w}= \\
&\mathbf{1}\left\{\operatorname{argmax}\left(\mathcal{C}\left(u_{o k}^{h, w}\right)\right) \neq \operatorname{argmax}\left(\mathcal{C}\left(u_{o l}^{h, w}\right)\right)\right\},
\end{split}
\end{equation}
where $ \ell_2\left(\mathbf{p}_{oi}^k, \mathbf{p}_{oi}^{l}\right)=\left\|\mathbf{p}_{oi}^k-\mathbf{p}_{oi}^{l}\right\|_2^2 $ calculates the square of the euclidean distance of the confidence maps of the overlapping regions. $W$ and $H$ represent the width and height of output maps. $ k $ and $ l$ represent the sequence number of spatial relationships among the patch group $\mathcal{G}_{u_i}$.

\noindent \textbf{Discussion.} $ \mathcal{L}_{BCC} $ uses an elegant $\ell_{2}$ for the distance measure between the anchor and positive samples, which ensures the high efficiency of our entire framework. It achieves a 1.7\% and 8.98\% performance improvement on Cityscapes and MoNuSeg after the joint $\mathcal{M}_{imp}^{h,w}$, respectively. (see Table~\ref{tab:table4} and the supplementary file.)

\subsection{Stage II: Dynamic Pseudo-Label Memory Bank (Strong Constraint)}
\label{sec:stage 2}
To increase the quality of training pseudo-labels and prevent noise from adversely affecting the model. Supported by CWC, we come up with the concept of the DPM $\mathcal{Q}$ for selecting reliable pseudo-labels and dynamically maintaining a high confidence and high consistency pseudo-label repository to undertake pseudo-label supervision.

\textbf{How to dynamically update?} Previous attempts \cite{sohn2020fixmatch} to evaluate the reliability of pseudo-labels mainly focus on pixel-level filtering methods, with the common strategy being to filter out low-confidence pixel information via manual or adaptive thresholding. However, filtered pixels often include complex
and vital information, which may increase the negative consequences of the long-tail effect. ST++ \cite{yang2022st++} presents a method for image-level selection by evaluating the stability of pseudo-labels at various training phases. It needs to calculate the model's classification differences for numerous phases, which is time-consuming. In our framework, we investigate an efficient algorithm for evaluating the reliability of pseudo-labels in order to promote rewarding updating of DPM $\mathcal{Q}$. Specifically, we reckon that \textbf{the greater the semantic consistency of pseudo-labels in overlapping regions including patches from different contexts, the greater the reliability of pseudo-labels}. Thus, for each unlabeled image $u_i$, we randomly crop and form a patch group $ \mathcal{G}_{u_ i}=\left\{\left(u_i^1, u_i^2, u_i^3, u_i^4\right)\right\}_{i=1}^N $ having four adjacent and overlapping patches of size $H_{o} \times W_{o}$. We utilize the model developed currently to evaluate the meanIOU of the predictions of the four patches’ overlapping regions as the reliability score $\mathcal{R}_i$.
\begin{equation}
\setlength\abovedisplayskip{2pt}
\setlength\belowdisplayskip{2pt}
\label{con:deqn_ex7}
\mathbb{C}_{oi}=\sum_{1 \leqslant k<l \leqslant 4} \text {ConfusionMatrix}\left(y_{o i}^{k *}, y_{o i}^{l *}\right),
\end{equation}
\begin{equation}
\setlength\abovedisplayskip{2pt}
\setlength\belowdisplayskip{2pt}
\label{con:deqn_ex8}
\mathcal{R}_i=\operatorname{meanIOU}\left(\mathbb{C}_{oi}\right),
\end{equation}
where $y_{oi}^{k*}=\operatorname{argmax}\left(\mathcal{C}\left(u_{oi}^k\right)\right)$ is pseudo masks of overlapping regions $u_{oi} $. After getting the reliability scores of all unlabeled images, we sort the entire set of unlabeled images based on these scores and select the Top-K\% reliable images pseudo-labels and corresponding images to entry into the DPM $ \mathcal{Q} $. (The original pseudo-labels in the DPM will be totally replaced.)

\textbf{When will an update be made?} To maintain the optimal pseudo-labels in the DPM at all times, the DPM will be automatically updated when the model achieves a gain of $\mathcal{G}$ on the validation set or when it reaches a predefined $\mathcal{K}$ epochs of training. This ensures that the model always obtains more rewarding information from the DPM.

We employ pseudo-labels $y_{ui}^*$ from the DPM to rigorously supervise the confidence map $\mathbf{p}_{ui}$ of the corresponding model output using cross-entropy loss:
\begin{equation}
\setlength\abovedisplayskip{2pt}
\setlength\belowdisplayskip{2pt}
\label{deqn_ex9}
\mathcal{L}_{DPM}=\frac{1}{\left|\mathcal{B}_u\right|} \sum\limits_{\mathbf{X} \in \mathcal{B}_u} \frac{1}{W \times H} \sum\limits_{i=0}^{W \times H}\left(\ell_{c e}\left(\mathbf{p}_{ui}, y_{ui}^*\right)\right),
\end{equation}
where $ W $ and $ H $ represent the width and height of unlabeled images, respectively.

\noindent \textbf{Discussion.} Instead of always utilizing all unlabeled images and corresponding pseudo-labels \cite{dupre2019improving}, our DPM dynamically filters out the less-reliable pseudo-labels based on CWC. In addition, our DPM updates information more frequently and flexibly in order to assist the model in learning more rewarding unlabeled data. The experimental results listed in Table~\ref{tab:table4} and Figure~\ref{fig:fig4} demonstrate its benefits.

\section{Experiments}
\label{sec:sec4}
\subsection{Setup}
\noindent \textbf{Datasets.} We evaluate our approach on three publicly available benchmarks to encompass various application scenarios such as urban street scenes semantic segmentation, histopathological tissue detection, and land cover classification, represented by Cityscapes \cite{cordts2016cityscapes}, MoNuSeg \cite{kumar2019multi}, and DeepGlobe \cite{demir2018deepglobe}. Cityscapes contains 2975 training images with fine-annotated labels of 19 semantic classes and 500 validation images. We compare our method with state-of-the-art methods under 1/30, 1/16, 1/8, and 1/4 partition protocols following ST++ \cite{yang2022st++} and CPS \cite{chen2021semi}. MoNuSeg was published by the multi-organ nuclei segmentation challenge \cite{kumar2019multi} and consists of 30, 7, and 14 histopathologic images ($ 1000\times1000$ pixels) for training, validation, and testing, respectively. Our and other methods are implemented under 1/30, 1/6, 1/3, and full supervision partition protocols. DeepGlobe contains 803 satellite images ($2448\times2448$ pixels) that are applied for land cover classification analysis in the field of remote sensing. Following \cite{chen2019collaborative}, we divide the images into the training set, validation set, and test set with 454, 207, and 142 images, respectively. Similarly, our and other methods are implemented under 1/16, 1/8, 1/4, and full supervision partition protocols.

\noindent \textbf{Evaluation.} We employ DeepLabv3+ \cite{chen2018encoder} with ResNet-50 \cite{he2016deep} that has been pre-trained on ImageNet \cite{deng2009imagenet} as our segmentation model to ensure a fair comparison with prior work. (Experiments based on DeepLabv3+ with ResNet-101 are displayed in the supplementary file.) We use the mean intersection-over-union (mIOU) metric to evaluate the segmentation performance of all datasets, following previous work. For a comprehensive evaluation of MoNuSeg, we also used Dice coefficient (DC), Jaccard coefficient (JC), Specificity (SP), and Sensitivity (SE), which are commonly used in biomedical segmentation. We report results on the 500 Cityscapes \texttt{val} set, the 14 MoNuSeg \texttt{test} set, and the 142 DeepGlobe \texttt{test} set using only single-scale testing and without any post-processing techniques.

\noindent \textbf{Implementation details.} The batch-size for all datasets with a single NVIDIA TITAN RTX GPU is set to 2 in Stage I and 4 in Stage II. The initial learning rate of Stage I of the backbone is 0.005, 0.004, and 0.005 for Cityscapes, MoNuSeg, and DeepGlobe, respectively, whereas the learning rate of the segmentation head is 10 times that of the backbone. The initial learning rate of Stage II is reduced to 0.003, 0.003, and 0.004. We use the SGD optimizer to train Cityscapes, MoNuSeg, and DeepGlobe for 280, 200, and 150 epochs under a poly learning rate scheduler, respectively. Following ST++ \cite{yang2022st++}, the labeled data is randomly flipped and resized between 0.5 and 2.0, and the unlabeled images are augmented with colorjitter, grayscale, and blur. Due to memory limitations, we train the model with a smaller crop size than CPS \cite{chen2021semi} (720 for Cityscapes and 512 for MoNuSeg and DeepGlobe). To get a fair comparison result for Cityscapes, we employ OHEM loss with the same parameters as previous work. For MoNuSeg and DeepGlobe, we train all models using only standard cross-entropy loss, without Sync-BN \cite{ioffe2015batch} and auxiliary loss. Moreover, the trade-off weights 
$\lambda_{BCC}$ and $\lambda_{DPM}$ are set to 0.16 and 1.0, respectively. And the selection of reliable pseudo-labels in Stage II selects, by default, the top 50\% of data for storage in the DPM. When the validation set metric achieves a 2\% gain or $count$ reaches 25, the DPM is automatically updated with more trustworthy pseudo-labels.

\subsection{Comparison with State-of-the-Art Methods}
An innovative framework based on the CWC traits of images is proposed. In this section, we implement advanced methods on various datasets with the same segmentation network and setting to ensure the fairness of comparison.

\begin{table}[]
\resizebox{85mm}{28mm}{
\begin{tabular}{c|c|cccc}
\toprule
Method        & Publication & \begin{tabular}[c]{@{}c@{}}1/30\\ (100)\end{tabular} & \begin{tabular}[c]{@{}c@{}}1/16\\ (186)\end{tabular} & \begin{tabular}[c]{@{}c@{}}1/8\\ (372)\end{tabular} & \begin{tabular}[c]{@{}c@{}}1/4\\ (744)\end{tabular} \\ \midrule
SupOnly$^\dagger$                 & -                                                & 55.1                                                                     & 61.8                                                                     & 66.2                                                                    & 72.3                                                                    \\
CPS \cite{chen2021semi}                     & CVPR'21                                          & -                                                                        & 69.8                                                                     & 74.4                                                                    & 76.9                                                  \\
CAC \cite{lai2021semi}                     & CVPR'21                                          & 60.9                                                                     & 69.4                                                                     & 74.0                                                                    & -                                                                       \\
DARS \cite{he2021re}                    & ICCV'21                                          & -                                                                        & 66.9                                                                     & 73.7                                                                    & -                                                                       \\
ST++$^\dagger$ \cite{yang2022st++}                  & CVPR'22                                          & 61.4                                                   & 70.1                                                                        & 73.2                                                                       & 74.7                                                                       \\
U$^2$PL \cite{wang2022semi}                 & CVPR'22                                          & 59.8                                                                     & -                                                                        & 73.0                                                                    & 76.3                                                                    \\
USRN \cite{guan2022unbiased}                    & CVPR'22                                          & -                                                                        & \textsl{\textbf{71.2}}                                                   & 75.0                                                 & -                                                                       \\
PS-MT \cite{liu2022perturbed}                   & CVPR'22                                          & -                                                                        & -                                                                        & 74.4                                                                    & 75.2                                                                    \\
CPCL \cite{10042237}                   & TIP'23                                          & -                                                                        & 69.9                                                                        & 74.6                                                                    & 77.0                                                                    \\
UniMatch \cite{unimatch}                   & CVPR'23                                          & \textsl{\textbf{64.5}}                                                                        & -                                                                        & \textsl{\textbf{75.6}}                                                                    & \textsl{\textbf{77.4}}                                                                    \\ \midrule
\textbf{Ours}           & -                                                & \textbf{67.3}                                                            & \textbf{72.8}                                                            & \textbf{76.6}                                                           & \textbf{77.6}                                                           \\ \bottomrule
\end{tabular}}
\vspace{-0.5em}
\caption{Comparison with state-of-the-art methods on \textbf{Cityscapes} \texttt{val} set under different partition protocols. We use DeepLabv3+ as the segmentation network and ResNet-50 as the backbone. “SupOnly” means supervised training without using any unlabeled data. $\dagger$ means we reproduce the approach and other results are collected from \cite{chen2021semi,guan2022unbiased,liu2022perturbed}. The best results are marked in \textbf{bold} and the second best ones are marked in \textsl{\textbf{italic bold}}.}
\vspace{-1.5em}
\label{tab:table1}
\end{table}

\noindent \textbf{Cityscapes.} Table~\ref{tab:table1} shows the results of our method compared with other state-of-the-art methods on the Cityscapes dataset. We reproduce the representative methods within the same network and setting according to their publicly available codes or use the results reported in the original papers. Our framework achieves a stable improvement under different partition protocols. Specifically, our framework outperforms the supervised baseline (SupOnly) by +12.2\%, +11.0\%, +10.4\%, and +5.3\% under 1/30, 1/16, 1/8, and 1/4 partition protocols, respectively. Besides, ours outperforms new state-of-the-art methods with larger margins by 2.8\% , 1.6\%, and 1.0\% in mIOU for 1/30, 1/16, and 1/8 split of Cityscapes. Additionally, the training burden is analyzed in the supplementary file.

\noindent \textbf{MoNuSeg.} Table~\ref{tab:table2} shows the comparison results on the MoNuSeg dataset. Compared with latest and advanced methods, ours surpasses them with large margins under all partition protocols (especially when there are few labeled samples participating in training). The DC gap between the SupOnly on full set (77.82\%) and our 1/30 labeled setting result (75.51\%) is only 2.31\%. Under the 1/6 labeled setting, ours outperforms the supervised baseline on the full set (79.62\% vs. 77.82\%). We also observe that even under full supervision, our framework still obtains a +3.18\% gain.

\begin{table}
\resizebox{85mm}{40mm}{
\begin{tabular}{c|c|ccccc}
\toprule
Partition                  & Method        & DC (\%)                 & mIOU (\%)               & JC (\%)                 & SP (\%)                 & SE (\%)                 \\ \midrule[1pt]
\multirow{4}{*}{1/30 (1)} & SupOnly       & 61.87                   & 60.13                   & 45.54                   & 79.23                   & \textsl{\textbf{82.32}} \\
                          & CutMix \cite{yun2019cutmix}        & 65.35                       & 64.89                       & 48.75                       & \textsl{\textbf{86.16}}                       & 75.32                       \\
                          & ST++ \cite{yang2022st++}          & \textsl{\textbf{68.68}} & \textsl{\textbf{67.05}} & \textsl{\textbf{52.49}} & 85.83 & \textsl{\textbf{81.30}} \\
                          & UniMatch \cite{unimatch}          & 63.77 & 63.51 & 47.15 & 85.22 & 75.42 \\
                          & \textbf{Ours} & \textbf{75.51}          & \textbf{74.46}          & \textbf{60.89}          & \textbf{92.06}          & 79.47                   \\ \midrule
\multirow{5}{*}{1/6 (5)}  & SupOnly       & 73.16                   & 71.95                   & 58.52                   & 89.20                   & \textsl{\textbf{84.53}}\\
                          & CutMix \cite{yun2019cutmix}         & 74.71                       & 73.77                       & 59.91                       & 90.62                       & 83.96                       \\
                          & CAC \cite{lai2021semi}           & 74.10                   & 73.40                   & 59.21                   & 91.72                   & 79.17                   \\
                          & ST++ \cite{yang2022st++}          & \textsl{\textbf{77.85}} & \textsl{\textbf{76.46}} & \textsl{\textbf{63.83}} & 92.37 & \textbf{84.60}          \\
                          & UniMatch \cite{unimatch}          & 74.96 & 75.32 & 60.43 & \textbf{97.48} & 66.54 \\
                          & \textbf{Ours} & \textbf{79.62}          & \textbf{78.62}          & \textbf{66.28}          & \textsl{\textbf{95.72}}          & 77.65                   \\ \midrule
\multirow{5}{*}{1/3 (10)}  & SupOnly       & 74.53                   & 72.72                   & 60.07                   & 88.38                   & \textsl{\textbf{87.89}} \\
                          & CutMix \cite{yun2019cutmix}         & 76.72                       & 76.07                       & 62.73                       & 92.62                       & 82.44                       \\
                          & CAC \cite{lai2021semi}           & 74.43                   & 74.80                   & 59.63                   & \textbf{94.19}          & 73.62                   \\
                          & ST++ \cite{yang2022st++}          & 78.41 & 77.27 & 64.62 & 93.72 & 81.76                   \\
                          & UniMatch \cite{unimatch}          & \textsl{\textbf{79.15}} & \textsl{\textbf{77.84}} & \textsl{\textbf{65.63}} & \textsl{\textbf{93.51}} & 83.72 \\
                          & \textbf{Ours} & \textbf{80.37}          & \textbf{78.48}          & \textbf{67.27}          & 92.35                   & \textbf{88.29}          \\ \midrule
\multirow{5}{*}{Full (30)} & SupOnly       & 77.82                   & 76.82                   & 64.20                   & 92.59                   & \textsl{\textbf{83.89}}          \\
                          & CutMix \cite{yun2019cutmix}         & 78.13                       & 77.03                       & 64.50                       & 92.40                       & \textbf{85.09}                       \\
                          & CAC \cite{lai2021semi}           & \textsl{\textbf{80.71}}                       & \textsl{\textbf{79.35}}                       & \textsl{\textbf{67.73}}                       & \textsl{\textbf{94.38}}                       & 83.72                       \\
                          & ST++ \cite{yang2022st++}          & 79.98 & 78.80 & 66.79 & 94.27 & 82.54                   \\
                          & UniMatch \cite{unimatch}          & 78.15 & 77.31 & 64.33 & 94.31 & 79.66 \\
                          & \textbf{Ours} & \textbf{81.00}          & \textbf{79.73}          & \textbf{68.18}          & \textbf{94.75}          & 83.04 \\ \bottomrule
\end{tabular}}
\vspace{-0.5em}
\caption{Comparison with state-of-the-art methods on \textbf{MoNuSeg} \texttt{test} set under different partition protocols. All methods are reproduced by us via DeepLabv3+ with ResNet-50 for a fair comparison.}
\vspace{-1.5em}
\label{tab:table2}
\end{table}

\noindent \textbf{DeepGlobe.} We show the comparison results for the DeepGlobe dataset in Table~\ref{tab:table3}. Ours brings significant and stable improvements compared to the SupOnly and other popular advanced methods under all partition protocols. Besides, the mIOU gap between the SupOnly on full set (68.64\%) and our 1/4 labeled setting result (68.51\%) is only 0.13\%.

\begin{table}[]
\resizebox{85mm}{17mm}{
\begin{tabular}{c|c|cccc}
\toprule
Method        & Publication & \begin{tabular}[c]{@{}c@{}}1/16\\ (28)\end{tabular} & \begin{tabular}[c]{@{}c@{}}1/8\\ (56)\end{tabular} & \begin{tabular}[c]{@{}c@{}}1/4\\ (113)\end{tabular} & \begin{tabular}[c]{@{}c@{}}Full*\\ (454)\end{tabular} \\ \midrule
SupOnly                 & -                            & 55.47                                               & 62.19                                              & \textsl{\textbf{66.82}}                                               & 68.64                                                \\
CutMix \cite{yun2019cutmix}                  & ICCV'19                      & 53.33                                                   & 61.46                                                  & 66.07                                                   & 66.59                                                    \\
CAC \cite{lai2021semi}                     & CVPR'21                      & \textsl{\textbf{56.47}}                                                   & \textsl{\textbf{62.23}}                                                  & 66.26                                                   & \textsl{\textbf{69.45}}                                                    \\
ST++ \cite{yang2022st++}                    & CVPR'22                      & 55.61                                               & 62.21                                              & 65.67                                                   & 69.25                                                   \\ \midrule
\textbf{Ours}           & \textbf{-}                   & \textbf{58.36}                                      & \textbf{65.17}                                     & \textbf{68.51}                                      & \textbf{70.01}                                       \\ \bottomrule
\end{tabular}}
\vspace{-0.5em}
\caption{Comparison with state-of-the-art methods on \textbf{DeepGlobe} \texttt{test} set under different partition protocols. All methods are reproduced by us via DeepLabv3+ with ResNet-50 for a fair comparison. * means only using the BCC supervision (Stage I), and discussions about the full data setting are presented in Section~\ref{sec:seca8}.}
\vspace{-1.5em}
\label{tab:table3}
\end{table}

\subsection{Ablation Studies}
The notable contributions of our framework are condensed into 1) BCC loss with the importance factor, 2) an efficient method for evaluating reliability of pseudo-labels, and 3) the DPM. We conduct our ablation studies with DeepLabv3+ and ResNet-50 on the 1/8 split of Cityscapes to verify the effectiveness of them.

\noindent \textbf{Effectiveness of the BCC loss.} In Table~\ref{tab:table4}, we show the outcomes of the naive consistency loss (Exp. I) and our proposed BCC loss $\mathcal{L}_{BCC}$ with the importance factor $\mathcal{M}_{imp}^{h,w}$ (Exp. II), demonstrating that the importance factor $\mathcal{M}_{imp}^{h,w}$ can provide a +1.7\% gain over the supervised baseline method (Exp. SupOnly) with only labeled data. And the performance of the model declines (-1.5\%) when $\mathcal{L}_{BCC}$ is not used in the first stage of training (Exp. VII). They indicate that BCC loss encourages overlapping regions to maintain prominent semantic consistency. More ablation studies about different overlap sizes and number of patch pairs involved are discussed in the supplementary file.

\begin{table}[]
\centering
\resizebox{85mm}{14mm}{
\begin{tabular}{c|cccccc|c}
\toprule
ID            & \begin{tabular}[c]{@{}c@{}}$\mathcal{L}_{BCC}$ \\ (w/o $\mathcal{M}_{imp}^{h,w}$ )\end{tabular} & \begin{tabular}[c]{@{}c@{}}$\mathcal{L}_{BCC}$ \\ (w/ $\mathcal{M}_{imp}^{h,w}$ )\end{tabular} & \begin{tabular}[c]{@{}c@{}}Random select \\ (50\%)\end{tabular} & \begin{tabular}[c]{@{}c@{}}Reliable select \\ (50\%)\end{tabular} & \begin{tabular}[c]{@{}c@{}}All unlabeled \\ data\end{tabular} & \begin{tabular}[c]{@{}c@{}}DPM\\ $\mathcal{Q}$\end{tabular} & mIOU(\%)     \\ \midrule
SupOnly       &           &          &                      &                        &                    &             & 66.2          \\
I             & \ding{52}         &          &                      &                        &                    &             & 66.5          \\
II            &           & \ding{52}        &                      &                        &                    &             & 67.9          \\
III           &           & \ding{52}        & \ding{52}                    &                        &                    &             & 71.7          \\
IV            &           & \ding{52}        &                      & \ding{52}                      &                    &             & 73.0          \\
V             &           & \ding{52}        &                      &                        & \ding{52}                  &             & 73.1          \\
VI            &           & \ding{52}        & \ding{52}                    &                        &                    & \ding{52}           & 73.8          \\
VII            &           &         &                     &\ding{52}                        &                    & \ding{52}           & 75.1          \\\midrule
\textbf{Ours} &           & \ding{52}        &                      & \ding{52}                      &                    & \ding{52}           & \textbf{76.6} \\ \bottomrule
\end{tabular}}
\vspace{-0.5em}
\caption{Ablation study on the effectiveness of various components in our framework. $\mathcal{L}_{BCC}$ : biased cross-window consistency loss, $\mathcal{M}_{imp}^{h,w}$ : importance factor, Random select (50\%) means selecting 50\% pseudo-labels to retrain randomly. Reliable select (50\%) means selecting top-50\% reliable pseudo-labels to retrain based on Eqs.~\eqref{con:deqn_ex7} and \eqref{con:deqn_ex8}. All unlabeled data means using all pseudo-labels to retrain directly. DPM: dynamic pseudo-label memory bank $\mathcal{Q}$.}
\vspace{-1.0em}
\label{tab:table4}
\end{table}

\begin{table}[]
\centering
\resizebox{85mm}{6mm}{
\begin{tabular}{l|ccc}
\toprule
Minimum overlap size & $\frac{H}{3}$ or $\frac{W}{3}$    & $\frac{\textit{\textbf{H}}}{\textbf{2}}$ or $\frac{\textit{\textbf{W}}}{\textbf{2}}$ (default)             & $\frac{2H}{3}$ or $\frac{2W}{3}$    \\ \midrule
mIOU (\%)            & 67.1 & \textbf{67.9} & 67.6 \\ \bottomrule
\end{tabular}}
\vspace{-0.5em}
\caption{Ablation study on different minimum overlap sizes. $H$ and $W$ represent the height and width of the input image.}
\vspace{-1.0em}
\label{tab:table5}
\end{table}

\begin{table}[]
\centering
\resizebox{60mm}{6mm}{
\begin{tabular}{c|cccc}
\toprule
$n$         & 0    & 1    & 3    & 6 (default)   \\ \midrule
mIOU (\%) & 66.2 & 66.6 & 66.5 & \textbf{67.9} \\ \bottomrule
\end{tabular}}
\vspace{-0.5em}
\caption{Ablation study on different number of patch pairs in each patch group.}
\vspace{-1.0em}
\label{tab:table6}
\end{table}

\begin{table}[]
\centering
\begin{tabular}{l|c}
\toprule
Method                                      & mIOU(\%)               \\ \midrule
Pixel-level filtering                          & 72.5                   \\
\textbf{Our image-level reliability selection} & \textit{\textbf{73.0}} \\
\textbf{Ours}                               & \textbf{76.6}          \\ \bottomrule
\end{tabular}
\vspace{-0.5em}
\caption{Effectiveness of the pseudo-label reliability evaluation. Performance analysis over pixel-level filtering strategy and our image-level reliability selection strategy.}
\vspace{-1.0em}
\label{tab:table7}
\end{table}

\noindent \textbf{Impact of parameters of the BCC loss.} For each unlabeled image $u_i$, four adjacent and overlapping patches are defined as a patch group in stage I. And the default minimum overlap size is defined as $\frac{H}{2}$ or $\frac{W}{2}$.  The size of the overlapping area determines the difference in the context information contained between adjacent patches. We conduct an ablation study on different minimum overlap sizes, as shown in Table~\ref{tab:table5}, which demonstrates that the most suitable size for $\mathcal{L}_{BCC}$ is $\frac{H}{2}$ or $\frac{W}{2}$.

we further consider $n$ patch pairs with overlapping areas in each patch group. In Table~\ref{tab:table6}, we find that when $n$ reaches the maximum value of 6, the result outperforms other counterparts, which proves that the wider cross-window information involved is beneficial for mining unlabeled data.

\noindent \textbf{Effectiveness of the pseudo-label reliability evaluation.} We verify it from two perspectives: (1) as shown in Experiments III and IV in Table~\ref{tab:table4}, the mIOU gap between random selection (50\%) and reliable selection (50\%) based on Eq.~\eqref{con:deqn_ex7} and \eqref{con:deqn_ex8} is 1.3\%, indicating that our pseudo-label reliability evaluation is effective. (2) We directly train the model with all unlabeled images and their pseudo-labels, and its performance is nearly identical to that of a model trained with only 50\% high-reliable pseudo-labels. This indicates our approach's capability to reduce noise interference in pseudo-labels, as we mentioned in Section~\ref{sec:stage 2}.

In addition, to further illustrate the advantages of this image-level reliability selection, we implement a comparison with the pixel-level filtering method. Specifically, the pixel-level filtering method ignores the pseudo-label information of the pixels with a maximum class confidence of less than 0.75 during the training phase. As shown in Table~\ref{tab:table7}, our image-level reliability selection is superior to the pixel-level filtering approach.

\begin{figure}
\centering
\includegraphics[scale=.32]{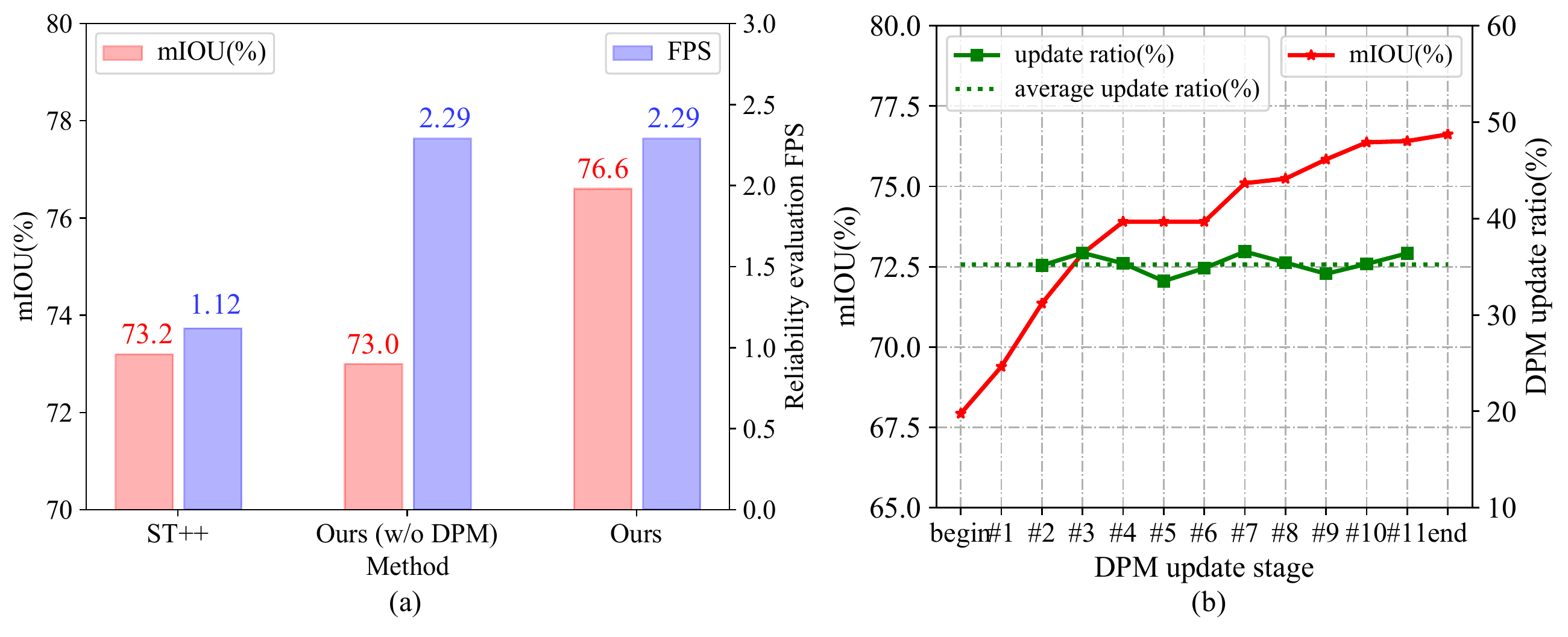}
\vspace{-2.5em}
\caption{(a) Comparison between the efficiency and accuracy of evaluating the reliability of pseudo-labels. (b) The model performance and the update ratio of the DPM at each automatic update stage. \#$k$: the $k$-th update and training. \texttt{begin}: the beginning of Stage II. \texttt{end}: the end of training.}
\label{fig:fig4}
\vspace{-1.0em}
\end{figure}

\noindent \textbf{Efficiency of the pseudo-label reliability evaluation.}
We employ the same settings and device to make a fair comparison with ST++ \cite{yang2022st++} on the efficiency of evaluating the reliability of pseudo-labels. ST++ needs to calculate the model’s
classification differences for numerous phases. In contrast to ST++ which needs to calculate model classification differences in multiple stages, our method only needs to calculate model differences in overlapping areas in a single stage. In Figure~\ref{fig:fig4}(a), the efficiency of the pseudo-label evaluation method based on CWC proposed by us \textbf{(2.29 FPS)} is twice that of ST++ \textbf{(1.12 FPS)}, which provides support for the high dynamic performance of DPM.

\noindent \textbf{Effectiveness of the DPM.} In Table~\ref{tab:table4}, the comparison between our method's result and that of Experiment IV demonstrates that the DPM can bring a significant improvement by +3.6\%. In addition, the difference between Experiment III and Experiment VI demonstrates the advantage of the DPM even when random selection is employed. 

Furthermore, Figure~\ref{fig:fig4}(b) illustrates the model performance of the DPM at each automatic update, as well as the renewal ratio of the images corresponding to the memory bank's pseudo-labels. Specifically, approximately 35\% of unlabeled images are changed at each update, demonstrating the high dynamics of the DPM. With the continual update of DPM, our framework's performance is also continuously enhanced, indicating that it is wise to gradually utilize high-reliability pseudo-labels instead of all pseudo-labels to optimize the model.

We also conduct ablation experiments on the ratio of selected reliable pseudo-labels. The default setting 50\% is effective enough, as demonstrated in Table~\ref{tab:table8}.

\begin{table}[]
\centering
\begin{tabular}{l|ccc}
\toprule
Ratio & 20\% & 50\%(default) & 80\% \\ \midrule
mIOU (\%)                & 74.5    & 76.6          & 76.3    \\ \bottomrule
\end{tabular}
\vspace{-0.5em}
\caption{Ablation study on the ratio of reliable pseudo-labels.}
\vspace{-1.5em}
\label{tab:table8}
\end{table}

\subsection{Qualitative results}
\begin{figure}
\centering
\includegraphics[scale=.31]{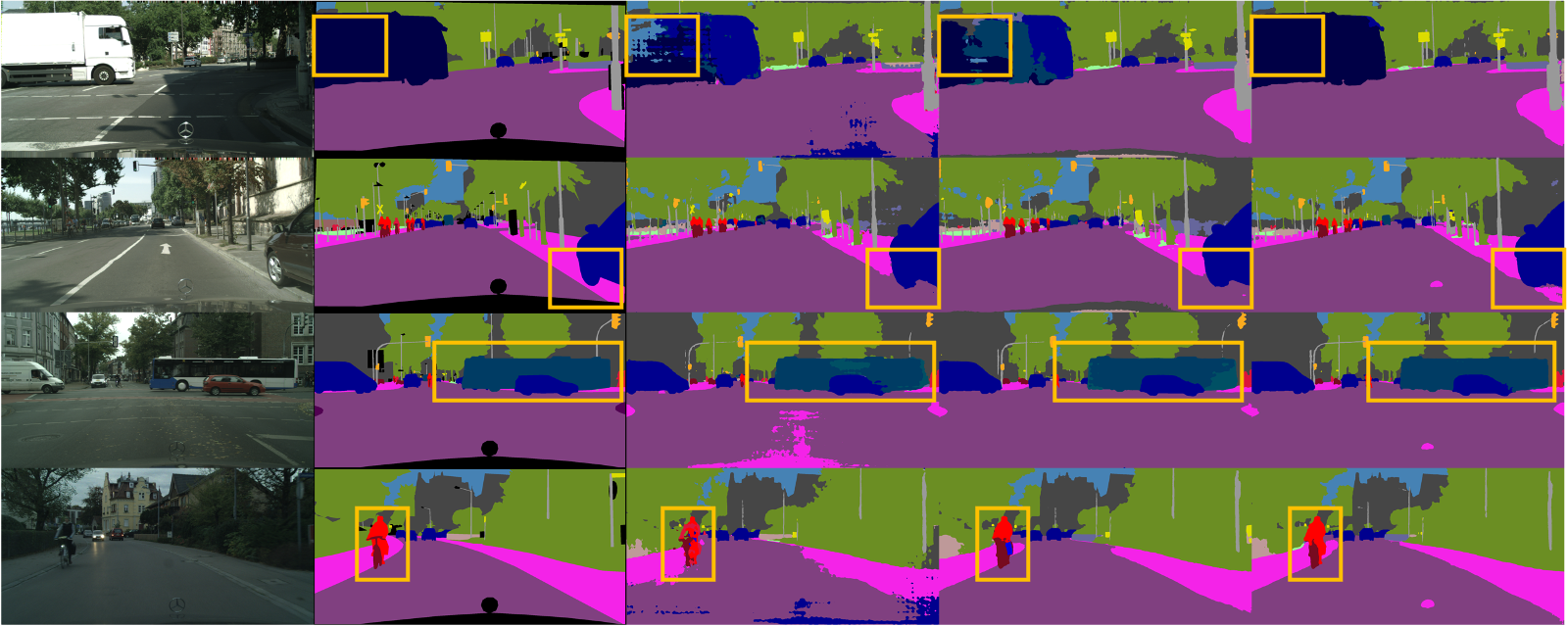}
\vspace{-2.0em}
\caption{Qualitative results on the Cityscapes set under the 1/8 protocol and ResNet-50 as the backbone. Columns from left to right denote
the original images, the ground-truth, the SupOnly results, the ST++ results, and our CWC results, respectively.}
\label{fig:fig5}
\vspace{-1.0em}
\end{figure}

Figure~\ref{fig:fig5} shows some segmentation results on Cityscapes. We can see that many mis-classified details like \texttt{truck}, \texttt{sidewalk}, and \texttt{person} in the SupOnly and ST++ results are corrected in CWC. More qualitative results on MoNuSeg and DeepGlobe datasets are displayed in the supplementary file.

\section{Discussion about the full data setting}
\label{sec:seca8}
In the full data setting on DeepGlobe dataset, images fed to the unsupervised branch are collected from the labeled training set, as CAC \cite{lai2021semi} does. Here we add an ablation study on the effect of the varying scale of unlabeled data in the full data setting. Table~\ref{tab:t8} shows that BCC supervision (Stage I) is also beneficial in the full data setting, and with the increase in the scale of unlabeled data, performance gradually improves. It is recommended to only use Stage I leads to the optimal performance in the full data setting.
\vspace{-1.0em}
\begin{table}[H]
\centering
\resizebox{72mm}{7mm}{
\begin{tabular}{cccccc}
\toprule
Scale    & 1/16 (28) & 1/8 (56) & 1/4 (113) & 1/2 (227) & Full (454) \\ \midrule
SupOnly  & -     & -   & -        & -     & 68.64      \\
Stage I  & 68.29     & 68.31    & 68.63        & 69.03     & \textbf{70.01 \textcolor{red}{(+1.37)}}      \\
Stage II & 68.13     & 67.37    & 66.86     & 66.11     & 65.47      \\ \bottomrule
\end{tabular}}
\vspace{-0.5em}
\caption{Results (mIOU\%) about the varying scale of unlabeled data in the full data setting on DeepGlobe \texttt{test} set.}
\label{tab:t8}
\end{table}

\section{Conclusion}
We propose a progressive learning framework for developing CWC systematically via mining weak-to-strong constraints. At the early stage, we propose a BCC loss with the importance factor to encourage the model to maintain consistency across overlapping confidence maps in different windows but does not restrict specific class attributes. We conceptualize the DPM to dynamically update and manage high-reliability pseudo-labels to strongly constrain the model in the latter period. The evaluation strategy of pseudo-label reliability based on CWC is the key of DPM. Our framework achieves the state-of-the-art performance across three representative datasets from various fields.  

\clearpage
\begin{appendix}
\renewcommand\thefigure{\Alph{section}\arabic{figure}} 
\renewcommand\thetable{\Alph{section}\arabic{table}}
\section{Comparison of different backbones}
\label{sec:seca2}
Similar to previous methods, we also adopt DeepLabv3+ \cite{chen2018encoder} with \textbf{ResNet-101} \cite{he2016deep} as the segmentation network and conducted experiments on Cityscapes dataset. Results in Table~\ref{tab:t1} demonstrate that our framework is still effective, without relying on a specific backbone.
\vspace{-1.0em}
\begin{table}[H]
\centering
\begin{tabular}{c|c|ccc}
\toprule
Method        & Publication & \begin{tabular}[c]{@{}c@{}}1/16\\ (186)\end{tabular} & \begin{tabular}[c]{@{}c@{}}1/8\\ (372)\end{tabular} & \begin{tabular}[c]{@{}c@{}}1/4\\ (744)\end{tabular} \\ \midrule
SupOnly$^\dagger$       & -           & 62.2                                                      & 69.1                                                    &  72.3                                                   \\
CCT \cite{ouali2020semi}           & CVPR'20     & 69.6                                                 & 74.5                                                & 76.4                                                \\
GCT \cite{ke2020guided}           & ECCV'20     & 66.9                                                 & 73.0                                                & 76.5                                                \\
CPS \cite{chen2021semi}           & CVPR'21     & \textit{\textbf{70.5}}                               & 75.7                             & 77.4                           \\
PS-MT \cite{liu2022perturbed}           & CVPR'22     & -                            & \textit{\textbf{76.9}}                              & 
\textit{\textbf{77.6}}                                       \\
ST++$^\dagger$ \cite{yang2022st++}           & CVPR'22     & 70.3                            & 73.9                              &  76.8                                      \\ \midrule
\textbf{Ours} & -           & \textbf{74.5}                                        & \textbf{77.0}                                           & \textbf{78.6}                                           \\ \bottomrule
\end{tabular}
\caption{Comparison with state-of-the-art methods on \textbf{Cityscapes} \texttt{val} set under different partition protocols. We use DeepLabv3+ as the segmentation network and \textbf{ResNet-101} as the backbone. “SupOnly” means supervised training without using any unlabeled data. $\dagger$ means we reproduce the approach and other results are collected from \cite{chen2021semi,liu2022perturbed}. The best results are marked in \textbf{bold} and the second best ones are marked in \textsl{\textbf{italic bold}}.}
\label{tab:t1}
\end{table}

\section{Ablation study on $\mathcal{M}_{imp}^{h,w}$}
\label{sec:seca3}
We conduct an ablation study that also consider pixels with large confidence difference (e.g., $>\tau$) in the same semantic classes ($\mathcal{M}$). Table~\ref{tab:t9} reveals that adding $\mathcal{M}$ has a negligible effect on the performance of the model.
\begin{table}[h]
\centering
\resizebox{55mm}{7mm}{
\begin{tabular}{cccc}
\toprule
$\tau$ & 0.4    & 0.5     & 0.6   \\ \midrule
$\mathcal{L_{BCC}}$ (w/ $\mathcal{M}_{imp}^{h,w}$)& \textbf{67.9} & \textbf{67.9} & \textbf{67.9}\\
$\mathcal{L_{BCC}}$ (w/ $\mathcal{M}_{imp}^{h,w}$ and $\mathcal{M}$) & \textbf{67.9}      & 67.6    & 67.7     \\ \bottomrule
\end{tabular}}
\caption{Results of ablation study on $\mathcal{M}$ on Cityscapes \texttt{val} set with 1/8 labeled data.}
\vspace{-0.5em}
\label{tab:t9}
\end{table}

\section{Analysis on MoNuSeg dataset}
\label{sec:seca5}
We further conduct analysis on the effectiveness of components of our framework on MoNuSeg dataset. As shown in Table~\ref{tab:t4}, the comparison between Exps. SupOnly and II demonstrates that the proposed BCC loss makes improvements by 8.98\%. In addition, we compare the performance of the BCC loss with and without the importance factor in Exps. I and II, which shows the contribution of the importance factor.

In light of the comparison among Exps. III, IV, and V, the advantage of the pseudo-label reliability evaluation method we proposed has been proved. Training with only the top-50\% of the reliable pseudo-labels (Exp. IV) can achieve more significant improvement (6.12\% and 3.87\%) than training with randomly selected equivalent pseudo-labels (Exp. III) or even all pseudo-labels (Exp. V).
\begin{table}[]
\centering
\resizebox{85mm}{14mm}{
\begin{tabular}{c|cccccc|c}
\toprule
ID            & \begin{tabular}[c]{@{}c@{}}$\mathcal{L}_{BCC}$ \\ (w/o $\mathcal{M}_{imp}^{h,w}$ )\end{tabular} & \begin{tabular}[c]{@{}c@{}}$\mathcal{L}_{BCC}$ \\ (w/ $\mathcal{M}_{imp}^{h,w}$ )\end{tabular} & \begin{tabular}[c]{@{}c@{}}Random select \\ (50\%)\end{tabular} & \begin{tabular}[c]{@{}c@{}}Reliable select \\ (50\%)\end{tabular} & \begin{tabular}[c]{@{}c@{}}All unlabeled \\ data\end{tabular} & \begin{tabular}[c]{@{}c@{}}DPM\\ $\mathcal{Q}$\end{tabular} & DC (\%) \\ \midrule
SupOnly       &           &          &                      &                        &                    &             & 61.87          \\
I             & \ding{52}         &          &                      &                        &                    &             & 67.29          \\
II            &           & \ding{52}        &                      &                        &                    &             & 70.85         \\
III           &           & \ding{52}        & \ding{52}                    &                        &                    &             & 69.51          \\
IV            &           & \ding{52}        &                      & \ding{52}                      &                    &             & \textbf{75.63}          \\
V             &           & \ding{52}        &                      &                        & \ding{52}                  &             &  71.76         \\
VI            &           & \ding{52}        & \ding{52}                    &                        &                    & \ding{52}           & 75.43          \\ \midrule
\textbf{Ours} &           & \ding{52}        &                      & \ding{52}                      &                    & \ding{52}           & \textit{\textbf{75.51}} \\ \bottomrule
\end{tabular}}
\caption{Ablation study on the effectiveness of various components in our framework. The experiment data is \textbf{MoNuSeg} dataset under 1/30 partition protocols. $\mathcal{L}_{BCC}$ : biased cross-window consistency loss, $\mathcal{M}_{imp}^{h,w}$ : importance factor, Random select (50\%) means selecting 50\% pseudo-labels to retrain randomly. Reliable select (50\%) means selecting top-50\% reliable pseudo-labels to retrain. All unlabeled data means using all pseudo-labels to retrain directly. DPM: dynamic pseudo-label memory bank $\mathcal{Q}$.}
\label{tab:t4}
\end{table}

\section{Training burden}
\label{sec:seca9}
Table~\ref{tab:t7} shows the training burden of our approach at different stages. And the time consumed for each update of DPM is 20.6 minutes on Cityscapes dataset with 1/8 labeled data.

\begin{table}[H]
\centering
\resizebox{80mm}{6mm}{
\begin{tabular}{c|ccc}
\toprule
     & Training time (each epoch) (s) & GPU Memory (M) & Params (M) \\ \midrule
Stage I  & 309.81         & 15483     & 40.475    \\
Stage II & 713.85          & 13827     & 40.475     \\ \bottomrule
\end{tabular}}
\caption{The training burden of our method on Cityscapes dataset with 1/8 labeled data.}
\label{tab:t7}
\end{table}

\section{Qualitative results}
\label{sec:seca6}
\begin{figure*}[t]
		\includegraphics[scale=.65]{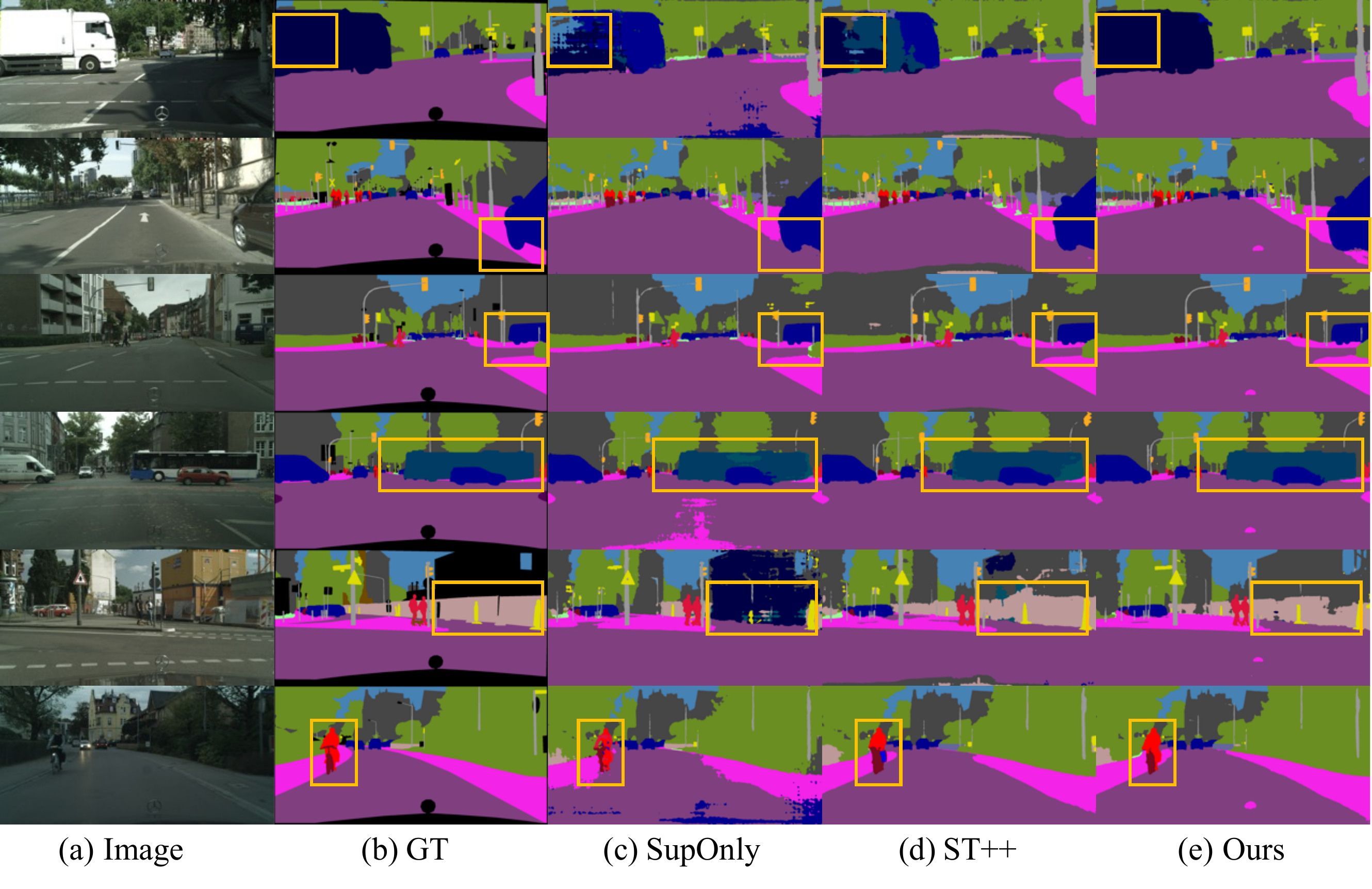}
	    \caption{Qualitative results on the Cityscapes val set.(a) and (b) are corresponding to images and Ground Truth(GT), (c) represents
the results of supervised baseline(SupOnly), (d) is the results of ST++ \cite{yang2022st++}, and (e) is the results of our framework. Orange rectangles highlight the difference among of them.}
	\label{fig:sup1}
\end{figure*}

\begin{figure*}[htbp]
		\includegraphics[scale=.56]{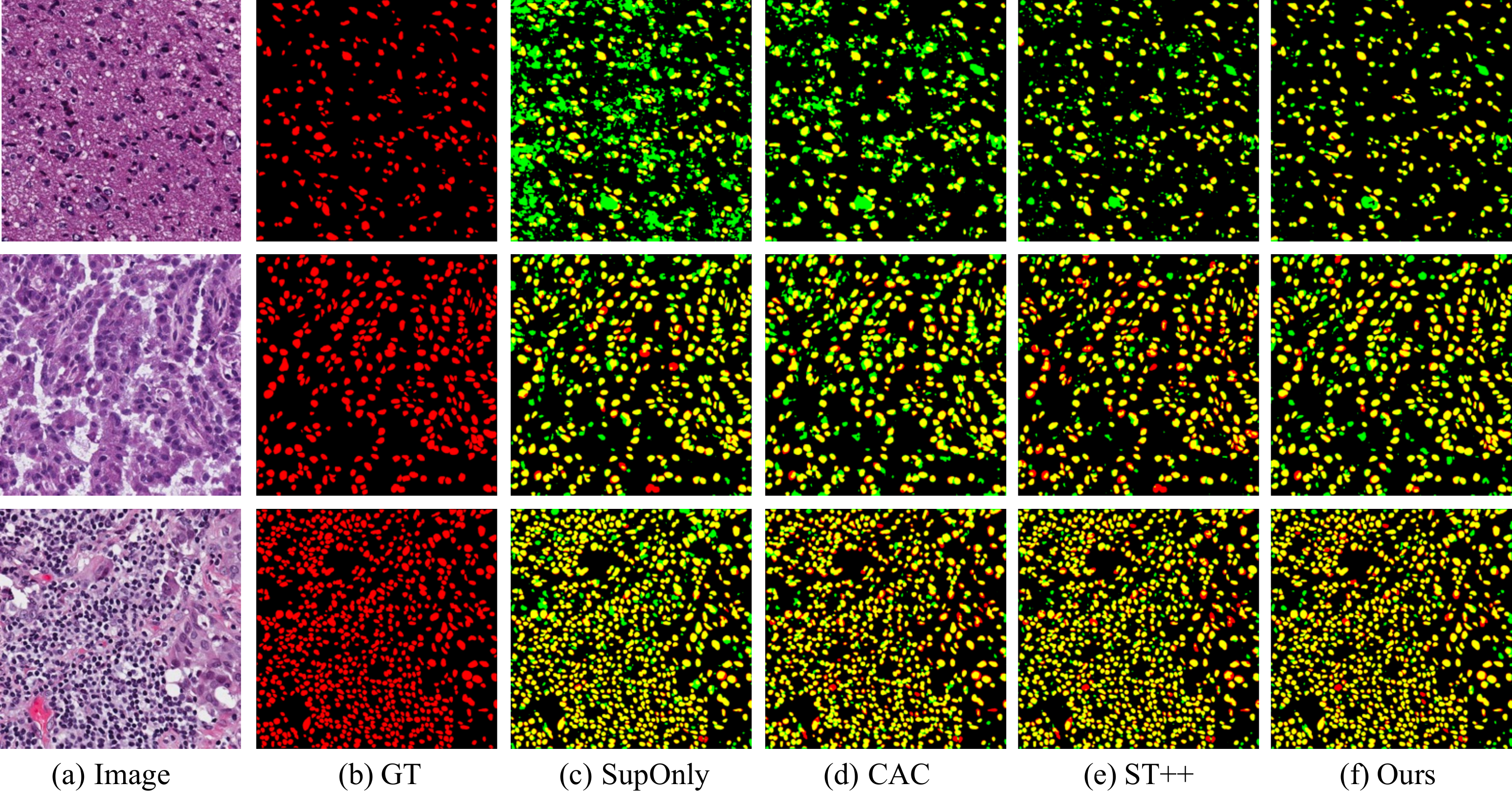}
	    \caption{Qualitative results on the MoNuSeg test set.(a) and (b) are corresponding to images and Ground Truth(GT), (c) represents
the results of supervised baseline(SupOnly), (d) and (e) is the results of CAC \cite{lai2021semi} and ST++ \cite{yang2022st++}, and (f) is the results of our framework. \textcolor{green}{Green} and \textcolor{red}{red} present the predictions and ground truth respectively, while \textcolor{yellow}{yellow} indicates their overlap regions.}
	\label{fig:sup2}
\end{figure*}

\begin{figure*}[htbp]
		\includegraphics[scale=.56]{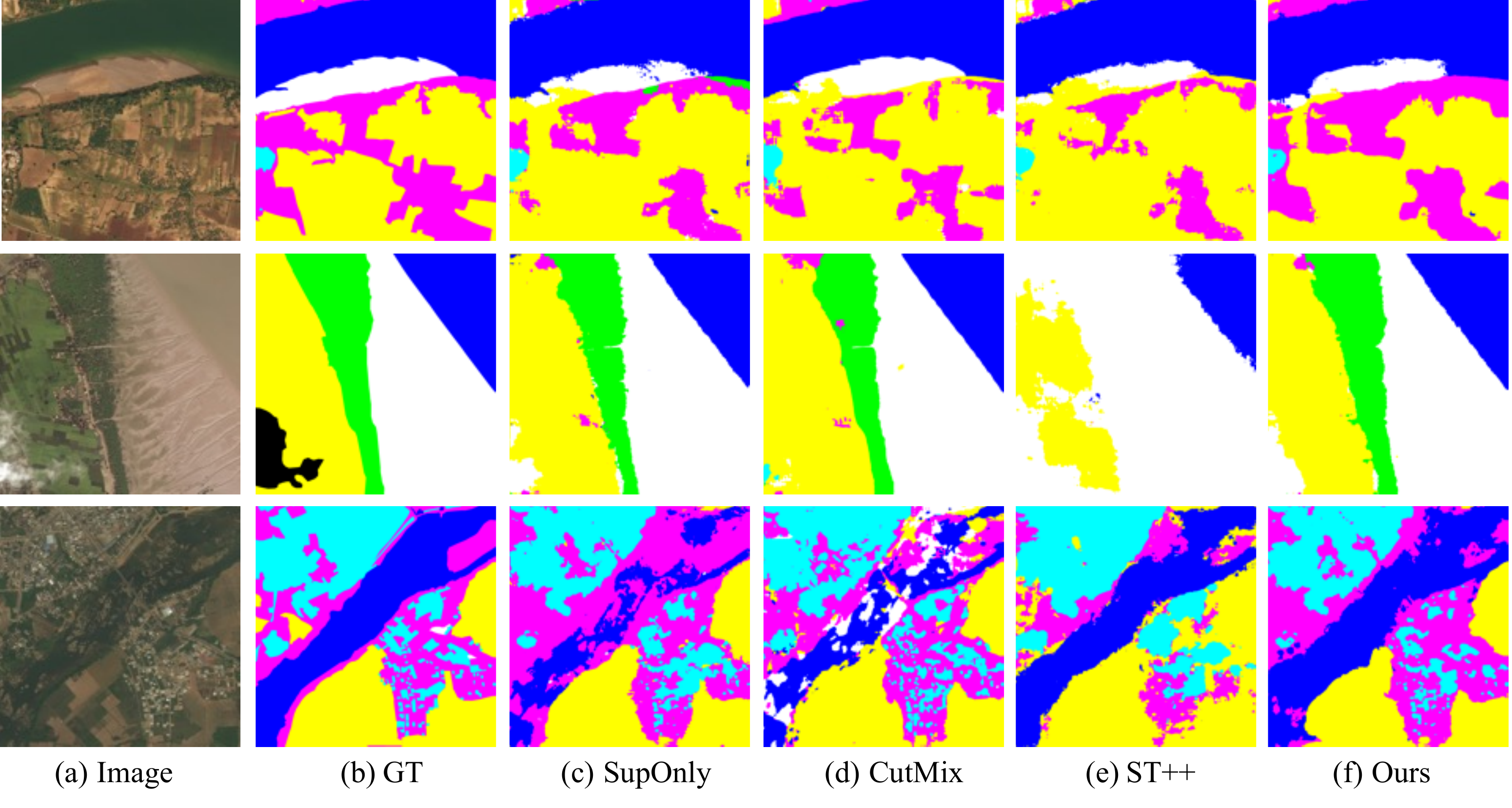}
	    \caption{Qualitative results on the DeepGlobe test set.(a) and (b) are corresponding to images and Ground Truth(GT), (c) represents
the results of supervised baseline(SupOnly), (d) and (e) is the results of CutMix \cite{yun2019cutmix} and ST++ \cite{yang2022st++}, and (f) is the results of our framework. \textcolor{cyan}{Cyan} represents “urban”, \textcolor{yellow}{yellow} represents “agriculture”, \textcolor{magenta}{magenta} represents “rangeland”, \textcolor{green}{green} represents “forest”, \textcolor{blue}{blue} represents “water”, white represents “barren” and black represents “unknown”.}
	\label{fig:sup3}
\end{figure*}

\noindent \textbf{Qualitative results on the Cityscapes val set.} On the Cityscapes val set, we present some qualitative results under 1/8 protocol in Figure~\ref{fig:sup1}. All of the methods are based on DeepLabv3+ with the ResNet-50. In comparison to the previous state-of-the-art method, our method displays more accurate segmentation results thanks to the proposed progressive learning strategy.

\noindent \textbf{Qualitative results on the MoNuSeg test set.} We present some qualitative results under 1/3 protocol in Figure~\ref{fig:sup2} on the MoNuSeg test set. All of the methods are based on DeepLabv3+ with the ResNet-50. 

\noindent \textbf{Qualitative results on the DeepGlobe test set.} We display some qualitative results under 1/4 protocol in Figure~\ref{fig:sup3} on the DeepGlobe test set. All of the methods are based on DeepLabv3+ with the ResNet-50. 

\section{Per-class results}
\label{sec:seca7}
In Tables~\ref{tab:t5} and \ref{tab:t6}, we present in detail the IOU performance of our results and some other methods for \textbf{per-class} on DeepGlobe and Cityscapes datasets, respectively. It is worth noting that our framework achieves the most significant improvement on the tailed classes (\eg~, '\texttt{wall}', '\texttt{rider}', '\texttt{truck}', '\texttt{bus}', and '\texttt{train}' ), indicating that our method alleviates the class imbalance issue to a certain extent.

\begin{table}[h]
\centering
\resizebox{85mm}{30mm}{
\begin{tabular}{cccccccc}
\toprule
\multicolumn{1}{c|}{}              & Urban & Agriculture & Rangeland & Forest & Water & \multicolumn{1}{c|}{Barren} & mIOU(\%) \\ \midrule
\multicolumn{8}{c}{Partition Protocol : 1/16 (28)}                                                                             \\ \midrule
\multicolumn{1}{c|}{SupOnly}       & 74.49      & 77.23            &  28.11         & 61.88       &  62.57     & \multicolumn{1}{c|}{28.53}       & 55.47    \\
\multicolumn{1}{c|}{Ours}          &  75.50     &  75.52           &  23.96         &   63.47     &  75.36     & \multicolumn{1}{c|}{36.36}       & 58.36    \\
\multicolumn{1}{c|}{\textit{Gain}} & \textcolor{blue}{+1.01}       & \textcolor{red}{-1.71}             & \textcolor{red}{-4.15}          & \textcolor{blue}{+1.59}       & \textcolor{blue}{+12.79}       & \multicolumn{1}{c|}{\textcolor{blue}{+7.83}}       & \textcolor{blue}{+2.89}    \\ \midrule
\multicolumn{8}{c}{Partition Protocol : 1/8 (56)}                                                                              \\ \midrule
\multicolumn{1}{c|}{SupOnly}       & 75.44      & 82.77            &  29.21         &  67.32      &  67.67     & \multicolumn{1}{c|}{50.73}       & 62.19    \\
\multicolumn{1}{c|}{Ours}          & 75.47      & 82.18            & 32.73          & 69.22       & 77.13      & \multicolumn{1}{c|}{54.32}       & 65.17    \\
\multicolumn{1}{c|}{\textit{Gain}} & \textcolor{blue}{+0.03}      & \textcolor{red}{-0.59}            & \textcolor{blue}{+3.52}          & \textcolor{blue}{+1.90}       & \textcolor{blue}{+9.46}      & \multicolumn{1}{c|}{\textcolor{blue}{+3.59}}       & \textcolor{blue}{+2.98}    \\ \midrule
\multicolumn{8}{c}{Partition Protocol : 1/4 (113)}                                                                             \\ \midrule
\multicolumn{1}{c|}{SupOnly}       & 72.56      &  85.09           & 33.17          & 76.99       & 74.15      & \multicolumn{1}{c|}{58.94}       & 66.82    \\
\multicolumn{1}{c|}{Ours}          &  77.37     &  84.80           &  35.36         &  76.36      &  77.18     & \multicolumn{1}{c|}{60.00}       & 68.51    \\
\multicolumn{1}{c|}{\textit{Gain}} & \textcolor{blue}{+4.81}      & \textcolor{red}{-0.29}            & \textcolor{blue}{+2.19}          & \textcolor{red}{-0.63}       & \textcolor{blue}{+3.03}      & \multicolumn{1}{c|}{\textcolor{blue}{+1.06}}       & \textcolor{blue}{+1.69}    \\ \midrule
\multicolumn{8}{c}{Partition Protocol : Full (454)}                                                                            \\ \midrule
\multicolumn{1}{c|}{SupOnly}       & 76.52      &  85.13           &  37.73         & 75.80       &  75.99     & \multicolumn{1}{c|}{60.68}       & 68.64    \\
\multicolumn{1}{c|}{Ours}          & 78.19      & 86.08            &  40.23         &  75.50      &  78.86     & \multicolumn{1}{c|}{61.16}       & 70.01    \\
\multicolumn{1}{c|}{\textit{Gain}} & \textcolor{blue}{+1.67}      & \textcolor{blue}{+0.95}            & \textcolor{blue}{+2.50}          & \textcolor{red}{-0.30}       & \textcolor{blue}{+2.87}      & \multicolumn{1}{c|}{\textcolor{blue}{+0.48}}       & \textcolor{blue}{+1.37}    \\ \bottomrule
\end{tabular}}
\caption{Results (IoU) of different classes on the \textbf{DeepGlobe} \texttt{test} set under different partition protocols. We use DeepLabv3+ as the segmentation network and ResNet-50 as the backbone. \emph{Gain}: The mIOU gain of between SupOnly and our approach.}
\label{tab:t5}
\end{table}

\begin{table*}[]
	\resizebox{\textwidth}{!}{
		\begin{tabular}{@{}c|ccccccccccccccccccc|c@{}}
			\toprule
			& \rotatebox{90}{road}                 & \rotatebox{90}{sidewalk}             & \rotatebox{90}{building}             & \rotatebox{90}{wall}                 & \rotatebox{90}{fence}                & \rotatebox{90}{pole}                  & \rotatebox{90}{light}                & \rotatebox{90}{sign}                 & \rotatebox{90}{vegetation}           & \rotatebox{90}{terrain}              & \rotatebox{90}{sky}                  & \rotatebox{90}{person}               & \rotatebox{90}{rider}                & \rotatebox{90}{car}                  & \rotatebox{90}{truck}                & \rotatebox{90}{bus}                  & \rotatebox{90}{train}                & \rotatebox{90}{motorcycle}           & \rotatebox{90}{bicycle}              & \rotatebox{90}{mIOU(\%)}                 \\ \midrule
			
			\multicolumn{20}{c}{Partition Protocol : 1/16 (186)}                                                                                                                                                                                   &       \\ \midrule
			SupOnly                                                   & 96.0  & 73.9     & 89.4     & 29.2  & 41.3  & 54.9  & 62.9  & 70.8  & 90.6       & 56.6    & 92.3  & 74.2   & 44.5  & 91.5  & 36.7  & 42.0 & 23.6   & 39.8       & 70.7    & 62.2   \\
			Ours                                                    & 97.7   & 82.6     & 91.2     & 49.0  & 55.8  & 59.7  & 68.5  & 77.7  & 92.0       & 60.0    & 94.0  & 80.2   & 57.5  & 94.6  & 73.1  & 76.9  & 68.5  & 60.7      & 75.1    & 74.5    \\
   \emph{Gain}                                                    & \textcolor{blue}{+1.7}  & \textcolor{blue}{+8.7}     & \textcolor{blue}{+1.8}     & \textcolor{blue}{+19.8}  & \textcolor{blue}{+14.5}  & \textcolor{blue}{+4.8}  & \textcolor{blue}{+5.6}  & \textcolor{blue}{+6.9}  & \textcolor{blue}{+1.4}       & \textcolor{blue}{+3.4}    & \textcolor{blue}{+1.7}  & \textcolor{blue}{+6.0}   & \textcolor{blue}{+13.0}  & \textcolor{blue}{+3.1}  & \textcolor{blue}{+36.4}   & \textcolor{blue}{+34.9}  & \textcolor{blue}{+44.9}  & \textcolor{blue}{+20.9}      & \textcolor{blue}{+4.4}    & \textcolor{blue}{+12.3} \\ \midrule
			\multicolumn{20}{c}{Partition Protocol : 1/8 (372)}                                                                                                                                                                                                  \\ \midrule
			SupOnly                                                   & 96.5   & 77.1      & 90.7     & 37.6   & 51.5   & 60.1  & 64.9  & 74.8  & 91.5      & 55.8    & 93.4  & 76.6   & 51.5  & 93.1  & 52.0  & 67.2 & 48.5   & 55.5       & 73.6    & 69.1   \\
			Ours                                                    & 97.9  & 83.9     & 92.3     & 58.7   & 60.4  & 63.5  & 70.8  & 79.2  & 92.3       & 60.9    & 94.7   & 81.9   & 62.1  & 95.1  & 76.1  & 79.2  & 71.8  & 64.8      & 76.9    & 77.0    \\
   \emph{Gain}                                                    & \textcolor{blue}{+1.4}  & \textcolor{blue}{+6.8}     & \textcolor{blue}{+1.6}     & \textcolor{blue}{+21.1}  & \textcolor{blue}{+8.9}  & \textcolor{blue}{+3.4}  & \textcolor{blue}{+5.9}  & \textcolor{blue}{+4.4}  &  \textcolor{blue}{+0.8}      & \textcolor{blue}{+5.1}     & \textcolor{blue}{+1.3} & \textcolor{blue}{+5.3}   & \textcolor{blue}{+10.6}  & \textcolor{blue}{+2.0}  & \textcolor{blue}{+24.1}  & \textcolor{blue}{+12.0}  & \textcolor{blue}{+23.3}   & \textcolor{blue}{+9.3}      & \textcolor{blue}{+3.3}    & \textcolor{blue}{+7.9} \\ 
   \midrule
			\multicolumn{20}{c}{Partition Protocol : 1/4 (744)}                                                                                                                                                                                                  \\ \midrule
			SupOnly                                                   & 97.5   & 81.4     & 91.2     & 37.6  & 55.8  & 63.4  & 68.7  & 77.1  & 91.6       & 57.8    & 93.8  & 79.1   & 57.1  & 93.4  & 60.7  & 73.7 & 55.6   & 63.0       & 75.0    & 72.3   \\
			Ours                                                    & 97.7  & 83.3     & 92.8     & 61.9  & 63.1  & 64.9  & 71.2  & 79.7  & 92.5       & 60.0    & 94.4  & 82.8   & 64.4  & 95.3  & 76.8  & 87.1  & 78.8  & 68.7      & 77.8    & 78.6    \\
   \emph{Gain}                                                    & \textcolor{blue}{+0.2}  & \textcolor{blue}{+1.9}     & \textcolor{blue}{+1.6}     & \textcolor{blue}{+24.3}  & \textcolor{blue}{+7.3}  & \textcolor{blue}{+1.5}  & \textcolor{blue}{+2.5}  & \textcolor{blue}{+2.6}  & \textcolor{blue}{+0.9}       & \textcolor{blue}{+2.2}    & \textcolor{blue}{+0.6} & \textcolor{blue}{+3.7}   & \textcolor{blue}{+7.3}  & \textcolor{blue}{+1.9}  & \textcolor{blue}{+16.1}  & \textcolor{blue}{+13.4}  & \textcolor{blue}{+23.2}  & \textcolor{blue}{+5.7}      & \textcolor{blue}{+2.8}    & \textcolor{blue}{+6.3} \\\bottomrule
		\end{tabular}
	}
 \caption{Results (IoU) of different classes on the \textbf{Cityscapes} \texttt{val} set under different partition protocols. We use DeepLabv3+ as the segmentation network and ResNet-101 as the backbone. \emph{Gain}: The mIOU gain of between SupOnly and our approach. our framework achieves the most significant improvement on the tailed classes (\eg~, '\texttt{wall}', '\texttt{rider}', '\texttt{truck}', '\texttt{bus}', and '\texttt{train}' ), indicating that our method alleviates the class imbalance issue to a certain extent.}
	\label{tab:t6}
\end{table*}
\end{appendix}

\hspace*{\fill} \\
\hspace*{\fill} \\
\hspace*{\fill} \\
\hspace*{\fill} \\
\hspace*{\fill}\\
\hspace*{\fill}\\
\hspace*{\fill}\\

{\small
\bibliographystyle{ieee_fullname}
\bibliography{main}
}

\end{document}